\pdfoutput=1

\documentclass[11pt]{article}

\usepackage[final]{acl}

\usepackage{times}
\usepackage{latexsym}
\usepackage{multirow}
\usepackage{booktabs}
\usepackage{makecell}
\usepackage[T1]{fontenc}
\usepackage{amsmath}
\usepackage{amssymb}
\usepackage[utf8]{inputenc}
\usepackage{microtype}
\usepackage{tabularx}
\usepackage{array}
\usepackage[table]{xcolor}
\usepackage{inconsolata}
\usepackage{graphicx}
\usepackage[table]{xcolor}
\usepackage{enumitem}
\usepackage{xcolor}
\usepackage[most]{tcolorbox}
\usepackage{placeins}

\newtcolorbox[auto counter]{promptcard}[3][]{
  enhanced, breakable,
  before upper={\linespread{1.18}\selectfont},
  colback=black!2!white, colframe=black!55!white,
  boxrule=0.5pt, arc=1pt,
  left=5pt, right=5pt, top=4pt, bottom=4pt,
  before skip=10pt, after skip=10pt,
  fonttitle=\small\bfseries, coltitle=white, colbacktitle=black!65!white,
  title={Prompt~\thetcbcounter\quad #2},
  label={#3}, #1}
\newcommand{\cardmeta}[2]{{\footnotesize\textbf{#1:}~#2\par}}
\newcommand{\cardpart}[2]{\vspace{5pt}{\footnotesize\textbf{#1} --- \textit{#2}\par}\vspace{1pt}{\color{black!40}\hrule height 0.3pt}\vspace{2.5pt}}
\newenvironment{pbody}{\par\begingroup\footnotesize\ttfamily\obeylines\frenchspacing\raggedright\hyphenpenalty=10000\exhyphenpenalty=10000}{\endgroup\par}

\newtcolorbox[auto counter]{examplecard}[3][]{
  enhanced, breakable,
  before upper={\linespread{1.18}\selectfont},
  colback=blue!2!white, colframe=blue!25!black!60,
  boxrule=0.5pt, arc=1pt, left=5pt, right=5pt, top=4pt, bottom=4pt,
  before skip=10pt, after skip=10pt,
  fonttitle=\small\bfseries, coltitle=white, colbacktitle=blue!25!black!80,
  title={Example~\thetcbcounter\quad #2}, label={#3}, #1}
\newcommand{\evtag}[1]{~{\scriptsize\color{black!55}[#1]}}
\newcommand{\dturn}[3]{{\footnotesize\textbf{#1}#2:~#3\par}\vspace{1.5pt}}
\newcommand{\elide}[1]{{\footnotesize\color{black!55}\textit{[\,\dots\ #1 \dots]}\par}\vspace{1.5pt}}
\newcommand{\exdiag}[1]{\vspace{2pt}{\footnotesize\textbf{Why it is hard} --- \textit{#1}\par}}

\title{\textsc{PersEvalBench}: Can LLM Evaluators Judge Contextually Appropriate Use of Persona Information?}

\author{
\textbf{Xinyi Liu\textsuperscript{1,2}}\thanks{Work done during an internship at Amazon.},
\textbf{Hooshang Nayyeri\textsuperscript{1}},
\textbf{Emine Yilmaz\textsuperscript{1,3}},\\
\textbf{Dilek Hakkani-Tur\textsuperscript{1,2}},
\textbf{Charith Peris\textsuperscript{1}},
\textbf{Hari Thadakamalla\textsuperscript{1}}
\\[0.5em]
\textsuperscript{1}Amazon \quad
\textsuperscript{2}University of Illinois Urbana-Champaign \quad
\textsuperscript{3}University College London
\\
\small{\texttt{
\{hooshang, perisc, thadakah\}@amazon.com
}}
\\
\small{\texttt{\{liu323, dilek\}@illinois.edu} \quad \texttt{eminey@amazon.co.uk}}
\\[0.25em]
\small{\textbf{Correspondence:} \texttt{liu323@illinois.edu}}
}

\begin{document}
\maketitle

\begin{abstract}
Large language model (LLM) evaluators are increasingly used to assess personalized assistants, yet whether they can reliably judge personalization has rarely been tested. We introduce \textsc{PersEvalBench}, a benchmark testing whether an evaluator can recognize appropriate and inappropriate uses of user history. It contains 403 multi-episode scenarios where user information should persist, transfer, or cease to apply. The benchmark covers nine controlled error subtypes across three families: under-personalization, over-personalization, and misapplied-state errors. Human audits confirm that generated errors match their intended types and that the prescribed uses of user information are reasonable.

Across five LLM evaluators, recognition of controlled personalization errors remains limited even with the full dialogue history. On the human-evaluated subset, under-personalization shows the largest gap: humans recognize 94.4\% of these errors, compared with 63.9\% for the strongest reported evaluator. Evaluators often recognize that a response is problematic but reverse the error direction, mistaking failure to follow an applicable preference (UNDER) for use of an inapplicable preference (OVER). Easier evidence access offers little help: highlighting the relevant history turns raises error recognition only slightly, suggesting that the remaining difficulty lies in judging what role that information should play in the response.

\end{abstract}

\section{Introduction}
\label{sec:intro}

\begin{figure}[!t]
\centering
\includegraphics[width=1\columnwidth]{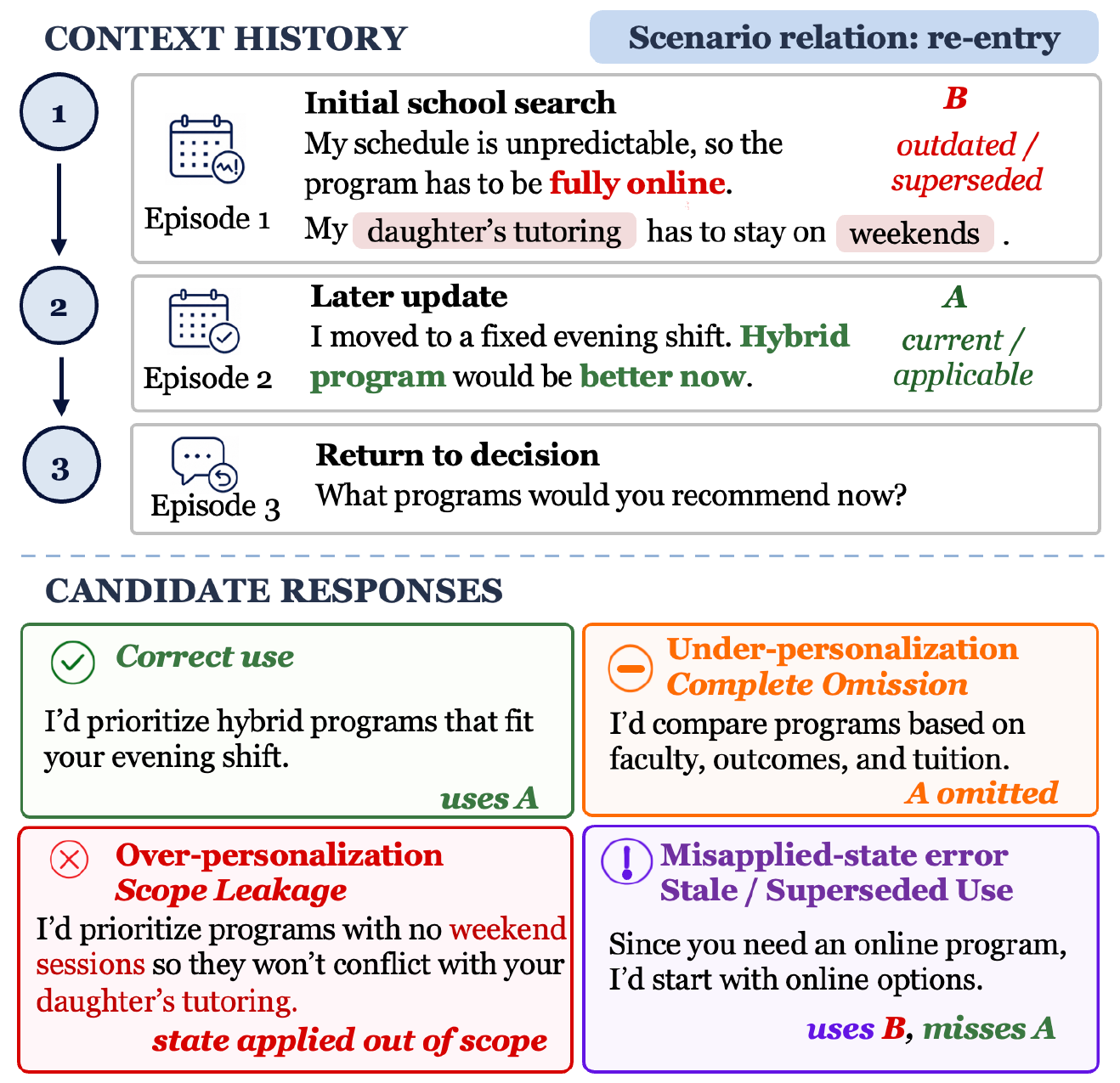}
\caption{Running example. An online-only requirement is superseded by a hybrid preference. Responses may use the update correctly, omit it, import an out-of-scope constraint, or continue using the stale state.}
\label{fig:example}
\end{figure}

\begin{figure*}[!t]
\centering
\includegraphics[width=\textwidth]{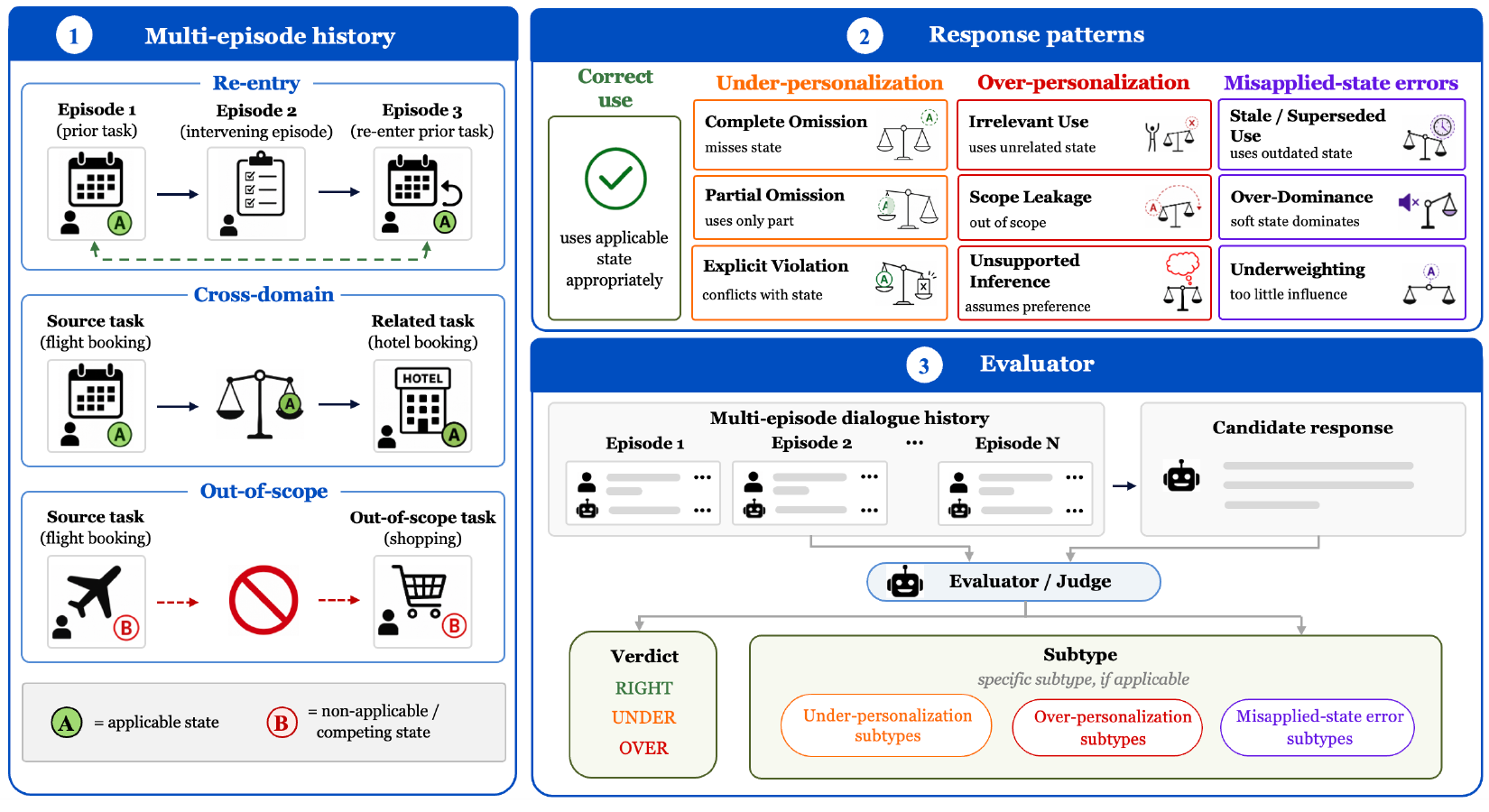}
\caption{Overview of \textsc{PersEvalBench}. Multi-episode histories determine whether prior user information should persist, transfer, or cease to apply. Each scenario pairs a \textsc{Right} response with one controlled error response, which the evaluator judges from the dialogue history.}
\label{fig:framework}
\end{figure*}

Personalization benchmarks increasingly rely on large language model (LLM) evaluators to score responses, compare systems, and guide model development. Existing work extensively tests whether assistants remember, follow, or over-apply information from user history \citep{maharana2024locomo, zhao2025prefeval, zhao2025personalens, xiao2026alpsbench, hu2026opbench, yoon2026benchpres, feng2026rpeval}, but rarely asks whether the evaluators themselves can judge these behaviors reliably. This judgment is both difficult and consequential: even factually correct information from history can be misused when it is outdated, belongs to someone else, falls outside the current scope, or is given the wrong weight, while evaluator errors can distort benchmark scores and system comparisons. Reliable personalization evaluation is therefore both necessary and challenging, requiring evaluators to judge not merely whether user information is used, but whether it is used appropriately in context.

Figure~\ref{fig:example} illustrates the challenge. A user changes an online-only requirement to a hybrid preference, so a response that still follows the old requirement uses real information from history that no longer applies. We introduce \textsc{PersEvalBench} to test whether evaluators can recognize such distinctions. Given a multi-episode history, a target request, and a candidate response, the evaluator predicts whether the response uses user information appropriately (\textsc{Right}), fails to use applicable information (\textsc{Under}), or uses information that should not apply (\textsc{Over}).

To make personalization judgments controlled and verifiable, we ground scenarios in Schema-Guided Dialogue (SGD) task structures \citep{rastogi2020towards} and define the intended use of user history before dialogue generation. Across three scenario relations, prior information is designed to persist, transfer to a related task, or cease to apply. Each scenario is then realized as a multi-episode dialogue with a matched \textsc{Right} response and controlled error response, keeping the task fixed while changing how user information is used. Nine error subtypes span under-personalization, over-personalization, and misapplied-state errors. Automated checks enforce the intended intervention during construction, and independent human audits validate the generated errors and the prescribed use of user information. The final benchmark contains 403 scenarios and 806 responses. Figure~\ref{fig:framework} summarizes the benchmark structure.

Across five LLM evaluators, performance remains limited even with the full dialogue history. The largest human-model gap appears in under-personalization. On the shared human-evaluation subset, human annotators recognize 94.4\% of these errors, while the strongest reported evaluator reaches 63.9\%. These errors are often misdiagnosed rather than simply missed: 26.6\% of under-personalized responses are labeled \textsc{Over}. Evaluators thus often treat a response that fails to follow applicable user information as if it had instead used information that should not apply. Easier evidence access does little to resolve this pattern, as highlighting the relevant history improves error recognition only slightly. This points to a deeper difficulty in judging how relevant user information should shape the response.

Our contributions are threefold. First, we define the evaluator-side task of judging whether a response uses information from user history appropriately, distinguishing under-personalization, over-personalization, and misapplied-state errors. Second, we introduce \textsc{PersEvalBench}, a controlled benchmark with nine error subtypes, matched \textsc{Right}/error responses, and human validation of both the generated errors and the prescribed use of user information. Third, we benchmark five LLM evaluators against human annotators and show that they remain unreliable, with particularly large failures on under-personalization and limited gains from easier evidence access.
\section{Related Work}
\label{sec:related}

\paragraph{Personalization and long-term memory benchmarks.}
Assistant-side personalization benchmarks primarily ask whether models can retain and follow user information across interactions. This includes persona-grounded dialogue \citep{zhang2018personalizing}, personalized generation \citep{salemi2024lamp}, long-term conversational memory \citep{maharana2024locomo}, preference following over long contexts \citep{zhao2025prefeval}, long-horizon preference following \citep{guo2026realpref}, alignment on real dialogues \citep{xiao2026alpsbench}, task-oriented personalization \citep{zhao2025personalens}, and dynamic or evolving user states \citep{jiang2025personamem,li2026horizonbench}. A complementary line examines whether models use stored information selectively, including over-personalization \citep{hu2026opbench}, rational preference utilization \citep{feng2026rpeval}, context-aware preference selectivity \citep{yoon2026benchpres}, and contextually appropriate memory use \citep{mireshghallah2025cimemories}, in the spirit of contextual integrity \citep{nissenbaum2004privacy}. These benchmarks evaluate the responding assistant, while the competence of the LLM judge used to assess it is generally assumed.


\newcommand{\yes}{\checkmark}
\newcommand{\no}{\textcolor{gray!60}{--}}

\begin{table}[t]
\centering
\caption{Task-level comparison. Contextual applicability asks whether prior information should affect the target response and with what influence. $\triangle$ denotes partial or mixed coverage.}
\label{tab:comparison}
\setlength{\tabcolsep}{4pt}
\renewcommand{\arraystretch}{1.08}

\resizebox{\columnwidth}{!}{
\begin{tabular}{@{}lcccc@{}}
\toprule
\shortstack[l]{\textbf{Benchmark}\\\textbf{family}}
& \textbf{Judge}
& \shortstack{\textbf{Multi}\\\textbf{episode}}
& \shortstack{\textbf{Contextual}\\\textbf{applic.}}
& \shortstack{\textbf{Controlled}\\\textbf{errors}} \\
\midrule

\shortstack[l]{Memory and\\preference following}
& \no
& \yes
& $\triangle$
& \no \\

\shortstack[l]{Selective\\personalization}
& \no
& $\triangle$
& \yes
& \no \\

Personalized Judge
& \yes
& \no
& \no
& \no \\

\addlinespace[1pt]
\rowcolor{gray!12}
\textbf{\textsc{PersEvalBench}}
& \yes
& \yes
& \yes
& \shortstack{\textbf{U/O/M}\\\textbf{9 subtypes}} \\

\bottomrule
\end{tabular}
}
\end{table}

\paragraph{LLM-as-judge and meta-evaluation.}
LLM evaluators are widely used for open-ended assessment \citep{zheng2023judging,liu2023geval}, with dedicated judge models trained for pairwise or fine-grained evaluation \citep{wang2024pandalm,kim2024prometheus,kim2024prometheus2,zhu2025judgelm}. Meta-evaluations test whether these judges agree with human assessments or recover controlled labels in instruction following \citep{zeng2024llmbar}, reward modeling \citep{lambert2024rewardbench}, and knowledge and reasoning tasks \citep{tan2025judgebench}. This line of work also documents systematic judge biases, including position and verbosity effects \citep{zheng2023judging,wang2024fair} and a preference for a judge's own generations \citep{panickssery2024selfpreference}. The matched pairs in \textsc{PersEvalBench} are constrained to comparable factual correctness, task relevance, completeness, specificity, and style (Section~\ref{sec:realization}), leaving these surface preferences little room to determine verdicts: what differs within a pair is the use of persona information. The most closely related evaluator-side work, Personalized Judge \citep{dong2024personalizedjudge}, tests whether an LLM can infer user preferences from persona descriptions. \textsc{PersEvalBench} instead tests whether an evaluator can judge a fixed response's use of information from a multi-episode history, using matched pairs that isolate under-personalization, over-personalization, and misapplied-state errors. Table~\ref{tab:comparison} summarizes this task-level distinction, and Table~\ref{tab:related-benchmarks} in Appendix~\ref{app:comparison} provides the full per-benchmark comparison.
\section{\textsc{PersEvalBench}}
\label{sec:method}

Figure~\ref{fig:framework} summarizes the benchmark structure: multi-episode histories establish how prior user information should apply, matched responses realize controlled personalization errors, and evaluators judge each candidate response from the dialogue history. We first define the evaluation task and controlled scenario specification, then describe how benchmark instances are constructed and human-verified.

\subsection{Task Definition}
\label{sec:task}

As shown in Figure~\ref{fig:framework} (3), each benchmark instance consists of a multi-episode dialogue history \(H\), a target user request \(q\), and a candidate assistant response \(r\). Given \((H,q,r)\), the evaluator assigns one of three verdicts based on how the response uses information from the user history: \textsc{Right}, \textsc{Under}, or \textsc{Over}. \textsc{Right} uses applicable information appropriately while avoiding information that should not affect the response. \textsc{Under} omits or gives too little influence to applicable information, whereas \textsc{Over} uses information that should not apply or gives it too much influence. For an error verdict, the evaluator also identifies one primary error subtype.

\subsection{Controlled Scenario Specification}
\label{sec:scenario-spec}

To make evaluator judgments controlled and interpretable, we determine the intended use of user history before generating the dialogue. Each specification fixes how prior information should affect the target response, what the correct response should do, and one controlled error to introduce. This separates the underlying personalization decision from its natural-language realization. Figure~\ref{fig:example} illustrates a re-entry scenario in which an earlier online-only constraint is superseded by an updated hybrid preference.

\paragraph{Scenario relations.}
\label{sec:scenario-grammar}
Figure~\ref{fig:framework} (1) illustrates the three history--target relations. We organize scenarios around the three broad ways prior information can relate to a target request in our task-oriented setting: it can remain applicable, transfer to a related task, or cease to apply. Accordingly, \emph{re-entry} tests whether an earlier state remains current, \emph{cross-domain transfer} whether it carries over and with what priority, and \emph{out-of-scope} whether seemingly relevant history should instead be ignored. Together, these relations cover the core applicability decisions of persistence, transfer, and exclusion.

\paragraph{Intended use of user information.}
We then make the intended influence of each user state explicit. Each state is assigned one of three roles for the target response: \textsc{Ignore}, \textsc{Support}, or \textsc{Dominate}, indicating that it should have no influence, may guide the response, or must be satisfied. The assignment depends on its holder, scope, temporal status, strength, and transferability to the target task. Each scenario includes an applicable state \(A\) and, when needed, a competing or non-applicable state \(B\). In Figure~\ref{fig:example}, the updated preference \(A\) is \textsc{Dominate}, while the superseded state \(B\) is \textsc{Ignore}. Appendix~\ref{app:state-relations} provides the complete inventory.

\paragraph{Error interventions.}
Figure~\ref{fig:framework} (2) summarizes the response patterns. Given the prescribed role of each user state, a response can depart from correct personalization in three basic ways: give applicable information too little influence, use information that should not apply, or assign influence to the wrong state or at the wrong level. These correspond to \emph{under-personalization}, \emph{over-personalization}, and \emph{misapplied-state} errors, respectively, and cover the core deviations from prescribed persona use in our setting. The taxonomy is also grounded in prior work on preference following \citep{zhao2025prefeval}, selective personalization \citep{hu2026opbench, yoon2026benchpres, feng2026rpeval}, and evolving user states \citep{jiang2025personamem, li2026horizonbench}.

To isolate each failure, every error response differs from its matched \textsc{Right} response by exactly one intervention. Under-personalization includes complete omission (CO), partial omission (PO), and explicit violation (EV); over-personalization includes irrelevant use (IU), scope leakage (SL), and unsupported inference (UI); and misapplied-state errors include stale or superseded use (SS), over-dominance (OD), and underweighting (UW).

Misapplied-state errors can support either error direction because they may simultaneously underuse one state and overuse another. We therefore count either \textsc{Under} or \textsc{Over} as recognition for this family. A human-authored compatibility matrix restricts interventions to scenario relations where they form natural violations, with subtype precedence defined in Appendix~\ref{app:grammar}; Appendix~\ref{app:examples} provides worked examples.

\paragraph{Evidence realization.}
Finally, we vary how easily the relevant user information can be recovered without changing the intended personalization decision. Holding the scenario relation, state roles, and error intervention fixed, we vary its evidence along two dimensions: form and distance. Evidence is \emph{explicit} when directly stated, \emph{implicit} when recoverable from a choice, rejection, or comparison, and \emph{accumulated} when distributed across multiple turns. We measure distance from the latest evidence turn to the target request as
\[
d_q = t_q - \max_{t \in \mathcal{E}_q} t,
\]
where \(\mathcal{E}_q\) denotes the evidence turns and \(t_q\) the target turn, and group instances into near (\(d_q \leq 3\)), medium (\(4 \leq d_q \leq 8\)), and far (\(d_q \geq 9\)) bins. Evidence form and distance vary how easily the evidence can be accessed rather than serving as benchmark targets; neither changes the prescribed use of user information or the benchmark label.

\subsection{Benchmark Realization}
\label{sec:realization}

We next turn each accepted specification into a natural dialogue and matched response pair, while preserving its predefined personalization logic.

\paragraph{Planning and verification.}
To obtain diverse but structured tasks, we expand the 20 domains and service schemas in the Schema-Guided Dialogue dataset \citep{rastogi2020towards} into 48 task types and a pool of 480 concrete task seeds. From this pool, an LLM planner instantiates 456 scenario specifications, each defining the episode goals, states \(A\) and \(B\), target request, candidate options, correct decision, error intervention, and evidence placement (Prompts~\ref{prompt:planner-sys} and~\ref{prompt:planner-user}).

Before dialogue generation, a verifier checks that each specification realizes the intended history--target relation, makes state applicability recoverable from the history, contains exactly one error intervention, and keeps the matched options comparable apart from persona use. Specifications that fail these checks are revised and re-evaluated, then discarded if they still do not satisfy the criteria (Prompts~\ref{prompt:verifier} and~\ref{prompt:refine}).

\paragraph{Dialogue synthesis.}
To separate scenario logic from its linguistic realization, a separate generator converts each accepted specification into a natural multi-episode dialogue without changing the predefined states, evidence, or target decision. The dialogue ends with the target request, and we record episode boundaries, the target turn, and evidence turns for each user state as benchmark metadata. These annotations support verification and the controlled context conditions in Section~\ref{sec:setup}, but are hidden from evaluators (Prompts~\ref{prompt:synth-sys} and~\ref{prompt:synth-user}).

\paragraph{Matched response generation.}
To isolate the assigned personalization error, we generate a matched response pair from each specification: one \textsc{Right} response and one error response. The \textsc{Right} response follows the prescribed state roles, while the error response receives only the information needed to realize its assigned intervention. Complete omission uses the target request alone, whereas other interventions receive subtype-specific restricted views of the relevant states. The two responses are constrained to remain comparable in factual correctness, task relevance, completeness, specificity, and style (Prompts~\ref{prompt:right} and~\ref{prompt:error}).

To make each intervention observable, scope leakage and unsupported inference must affect the recommendation, while underweighting must visibly reduce the influence of an applicable state. An automated validator checks observability, intervention fidelity, and pair comparability, regenerating failed responses with targeted feedback for up to three attempts (Prompt~\ref{prompt:validator}). Section~\ref{sec:verification} reports an independent human audit of the resulting responses.

\begin{table}[t]
\centering
\caption{Scale and composition of \textsc{PersEvalBench}.}
\label{tab:benchmark-summary}
\footnotesize
\setlength{\tabcolsep}{5pt}
\renewcommand{\arraystretch}{1.08}

\begin{tabular*}{0.92\columnwidth}{@{\extracolsep{\fill}}lr@{}}
\toprule
\multicolumn{2}{@{}l}{\textbf{Scale}} \\
\hspace{0.5em}Scenarios & 403 \\
\hspace{0.5em}Candidate responses & 806 \\
\hspace{0.5em}Avg.\ episodes / turns per dialogue & 2.4 / 26.8 \\

\addlinespace[3pt]
\multicolumn{2}{@{}l}{\textbf{Scenario relation}} \\
\hspace{0.5em}Re-entry & 174 (43.2\%) \\
\hspace{0.5em}Cross-domain transfer & 166 (41.2\%) \\
\hspace{0.5em}Out-of-scope & 63 (15.6\%) \\

\addlinespace[3pt]
\multicolumn{2}{@{}l}{\textbf{Error family}} \\
\hspace{0.5em}Under-personalization & 130 (32.3\%) \\
\hspace{0.5em}Over-personalization & 151 (37.5\%) \\
\hspace{0.5em}Misapplied-state & 122 (30.3\%) \\
\bottomrule
\end{tabular*}
\end{table}
\begin{table*}[t]
\centering
\caption{Error Recognition Rate (\%) by error subtype in the main setting, with FAR and the constant-\textsc{Over} baseline. Subtype abbreviations follow Section~\ref{sec:scenario-spec}. Und., Ov., and Mis.\ pool all error responses within each family, and All pools all error responses. Bold marks the best reported evaluator per column.}
\label{tab:main-err}
\footnotesize
\setlength{\tabcolsep}{3.8pt}
\resizebox{\textwidth}{!}{%
\begin{tabular}{@{}lccc!{\hspace{2.5pt}}c!{\hspace{2.5pt}}ccc!{\hspace{2.5pt}}c!{\hspace{2.5pt}}ccc!{\hspace{2.5pt}}c!{\hspace{2.5pt}}cc@{}}
\toprule
Model & \multicolumn{3}{c}{Under-personalization} &   & \multicolumn{3}{c}{Over-personalization} &   & \multicolumn{3}{c}{Misapplied-state} &   &   &   \\
\cmidrule(lr){2-4}\cmidrule(lr){6-8}\cmidrule(lr){10-12}
 & CO & PO & EV & Und. & IU & SL & UI & Ov. & SS & OD & UW & Mis. & All & FAR $\downarrow$ \\
\midrule
Kimi K2.5 & 63.6 & 48.7 & 44.7 & 52.3 & \textbf{78.1} & \textbf{70.5} & \textbf{90.7} & \textbf{79.5} & 61.3 & \textbf{42.6} & \textbf{93.2} & \textbf{65.6} & \textbf{66.5} & 24.6 \\
GLM-5 & 50.0 & \textbf{61.5} & 53.2 & 54.6 & 76.6 & 63.6 & 86.0 & 75.5 & 74.2 & 27.7 & 86.4 & 60.7 & 64.3 & 12.2 \\
DeepSeek-V3.2 & \textbf{65.9} & 51.3 & \textbf{57.4} & \textbf{58.5} & 62.5 & 50.0 & 86.0 & 65.6 & \textbf{77.4} & 25.5 & \textbf{93.2} & 63.1 & 62.5 & 10.9 \\
Qwen3-32B & 15.9 & 17.9 & 23.4 & 19.2 & 35.9 & 29.5 & 62.8 & 41.7 & 25.8 & 10.6 & 63.6 & 33.6 & 32.0 & \textbf{2.5} \\
\midrule
\rowcolor{gray!5}
Constant-\textsc{Over} & 0.0 & 0.0 & 0.0 & 0.0 & 100.0 & 100.0 & 100.0 & 100.0 & 100.0 & 100.0 & 100.0 & 100.0 & 67.7 & 100.0 \\
\bottomrule
\end{tabular}
}
\end{table*}

\paragraph{Coverage.}
We prioritize coverage of factors that determine contextual applicability rather than exhaustively crossing all design variables. Each valid relation--intervention combination is instantiated across multiple task types, goals, state configurations, and response decisions, with varied evidence form and distance. Appendix~\ref{app:grammar} reports the valid combinations, and Appendix~\ref{app:alignment} the final benchmark composition.

\subsection{Human Verification}
\label{sec:verification}

We conduct three complementary human studies on the same stratified sample of 100 scenarios. They test whether the generated errors realize their intended interventions, whether the prescribed use of user information is reasonable, and how well humans perform under the same information available to evaluators. Study~1 uses one annotator pair, while Studies~2 and~3 use a second, disjoint pair.

\paragraph{Intervention fidelity.}
Annotators independently judge whether each raw error response realizes its assigned intervention, followed by adjudication. The raw pipeline achieves 88\% subtype fidelity. We revise the 12 flagged responses while preserving the dialogue history and matched \textsc{Right} response, then re-verify them before inclusion in the benchmark. All subsequent human and model evaluations use this corrected benchmark.

\paragraph{State-role validation.}
To validate the prescribed use of user information independently of the construction pipeline, annotators assign each user state an \textsc{Ignore}, \textsc{Support}, or \textsc{Dominate} role without seeing the benchmark assignments. After adjudication, 86.3\% of these assignments match the benchmark, with substantial three-level agreement between annotators (weighted \(\kappa=.709\)); Appendix~\ref{app:human} provides the full analysis.

\paragraph{Human reference.}
Before seeing the benchmark policy, the same annotators judge candidate responses under the evaluator setting: they view the dialogue and each candidate response individually, without construction metadata, and assign \textsc{Right}, \textsc{Under}, or \textsc{Over}. Agreement over the 200 judgments is 83.5\% (Cohen's \(\kappa=0.72\)). This provides a human reference under the same information available to model evaluators (Appendix~\ref{app:human}).
\section{Experimental Setup}
\label{sec:setup}

\paragraph{Benchmark.}
Table~\ref{tab:benchmark-summary} summarizes the scale and composition of \textsc{PersEvalBench}. Each retained scenario pairs one \textsc{Right} response with one controlled error response. Of the 456 instantiated scenario specifications, 403 were retained after verification and quality control. The construction pipeline uses Claude Sonnet 4.6 \citep{anthropic2026sonnet46} for planning, specification verification, dialogue synthesis, response generation, and automated validation. Appendices~\ref{app:grammar} and~\ref{app:alignment} provide additional benchmark statistics and construction details.

\paragraph{Evaluators.}
We evaluate Claude Opus 4.6 \citep{anthropic2026opus46}, Kimi K2.5 \citep{kimiteam2026kimi25}, GLM-5 \citep{glmteam2026glm5}, DeepSeek-V3.2 \citep{deepseek2025v32}, and Qwen3-32B \citep{yang2025qwen3}.\footnote{Evaluator-specific results are partially omitted from the public version due to release restrictions.}

\paragraph{Evaluation settings.}
We vary three aspects of evaluation. \textbf{Context:} \emph{Target Only}, \emph{Recent}, and \emph{Full} provide the target request alone, the preceding six turns, and the complete history up to the target, respectively; \emph{Full + Gold Highlights} marks the annotated evidence turns within the full history, while \emph{Gold Evidence Only} provides only those turns and the target request in chronological order. \textbf{Formulation:} \emph{End-to-end} directly predicts the verdict and, when applicable, the subtype; \emph{Evidence-first} identifies relevant user states and evidence before predicting the verdict; and \emph{State-table} first assesses state applicability in a structured table. Unless otherwise stated, we use \emph{Full} with \emph{End-to-end} (Prompts~\ref{prompt:e2e}--\ref{prompt:agentic}). \textbf{Protocol:} we additionally evaluate pointwise scoring and direct pairwise comparison over matched responses (Appendix~\ref{app:protocols}).

\paragraph{Metrics.}
We compute Error Recognition Rate (ERR) over error responses and False Alarm Rate (FAR) over their matched \textsc{Right} responses. An error response counts as recognized when an under-personalization error is assigned \textsc{Under}, an over-personalization error is assigned \textsc{Over}, or a misapplied-state error is assigned either error verdict. The last case accepts both directions because the response underuses one state while overusing another. FAR is the percentage of \textsc{Right} responses incorrectly assigned \textsc{Under} or \textsc{Over}. Both metrics score the three-way verdict only; the elicited subtype label is not scored. We report both metrics because ERR alone can be inflated by an evaluator that predicts an error indiscriminately. For analyses over matched responses, we additionally report pair accuracy, which requires the evaluator to accept the \textsc{Right} response and correctly recognize its paired error response.

\section{Results}
\label{sec:results}

We ask three questions: how reliably LLM evaluators judge personalization and where they fail; whether easier evidence access reduces these failures; and whether alternative prompting or judging protocols improve their judgments.

\subsection{Overall Reliability and the Under-Personalization Failure}
\label{sec:results-main}

Table~\ref{tab:main-err} reports subtype-level error recognition and false alarms under the main full-history evaluation setting. The evaluated judges remain unreliable overall. A constant-\textsc{Over} baseline attains 67.7\% ERR but falsely flags every \textsc{Right} response. Error recognition must therefore be interpreted together with false alarms.

\begin{table}[!t]
\centering
\caption{Human and model performance on the same 100 scenarios under \textbf{Full} context. Each response is evaluated individually.}
\label{tab:human-main}
\footnotesize
\setlength{\tabcolsep}{4.2pt}
\renewcommand{\arraystretch}{1.08}
\begin{tabular}{@{}lccccc@{}}
\toprule
\textbf{Judge}
& \shortstack{\textbf{Und.}\\{\scriptsize \(n{=}36\)}}
& \shortstack{\textbf{Ov.}\\{\scriptsize \(n{=}40\)}}
& \shortstack{\textbf{Mis.}\\{\scriptsize \(n{=}24\)}}
& \textbf{All}
& \textbf{FAR $\downarrow$} \\
\midrule
Kimi K2.5 & 63.9 & \textbf{70.0} & 70.8 & 68.0 & 26.0 \\
GLM-5 & 58.3 & 60.0 & 54.2 & 58.0 & 9.0 \\
DeepSeek-V3.2 & 58.3 & 52.5 & 54.2 & 55.0 & 11.0 \\
Qwen3-32B & 25.0 & 32.5 & 29.2 & 29.0 & \textbf{1.0} \\
\midrule
Annotator A & \textbf{94.4} & 62.5 & \textbf{83.3} & \textbf{79.0} & 4.0 \\
Annotator B & \textbf{94.4} & 50.0 & 79.2 & 73.0 & 7.0 \\
\midrule
\textbf{Human mean} & 94.4 & 56.3 & 81.3 & 76.0 & 5.5 \\
\bottomrule
\end{tabular}

\end{table}

\textbf{Under-personalization.}
This family shows the largest human-model gap. On the shared 100-scenario subset, both human annotators recognize 94.4\% of under-personalization errors, compared with 63.9\% for the strongest reported evaluator (Table~\ref{tab:human-main}). Humans significantly outperform each evaluator in Table~\ref{tab:human-main} on these cases (Appendix~\ref{app:human}).

\begin{figure}[t]
\centering
\includegraphics[width=\columnwidth]{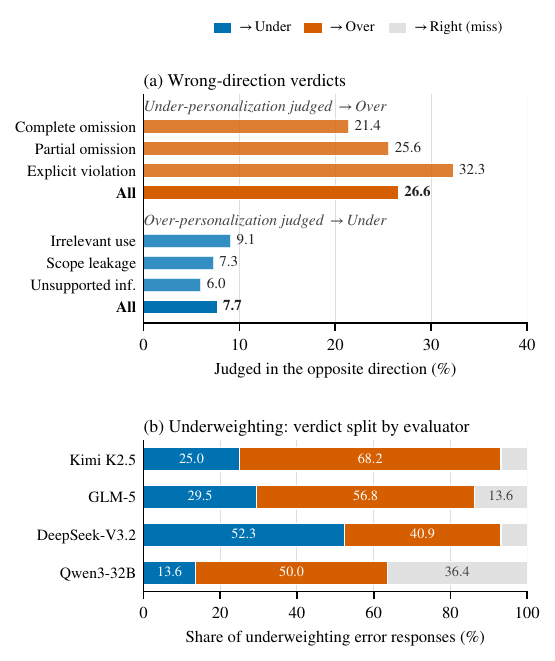}
\caption{Verdict behavior in the main setting. (a) Wrong-direction verdicts, pooled over evaluators: the share of under-personalization errors judged \textsc{Over} and of over-personalization errors judged \textsc{Under}. (b) Verdict split by reported evaluator for underweighting errors.}
\label{fig:verdicts}
\end{figure}

The dominant failure within this family is direction confusion. Evaluators assign 26.6\% of these responses the opposite \textsc{Over} verdict, far more often than over-personalization errors are judged \textsc{Under} (Figure~\ref{fig:verdicts}a). Secondary errors amplify but do not explain this pattern: on subsets that exclude human-flagged secondary \textsc{Over} errors, the rate remains 19.5--21.1\%, and the wrong-direction rationales ground their decisions in genuine history evidence but assign that evidence the wrong applicability, scope, or weight (Appendix~\ref{app:response_local}). The failure lies in determining its proper role in the response.

\textbf{Underweighting.}
Underweighting illustrates why misapplied-state errors admit both error directions. Evaluators split their error verdicts on these responses between \textsc{Under} and \textsc{Over} (Figure~\ref{fig:verdicts}b). The two verdicts capture complementary aspects of the same error: insufficient influence from the applicable state and excessive relative influence from a competing consideration.

\textbf{Evaluator operating points.}
The evaluators occupy distinct operating regimes rather than forming a single performance ranking. Kimi K2.5 is comparatively permissive, pairing 66.5\% ERR with the highest reported FAR of 24.6\%, whereas Qwen3-32B is strongly conservative, reducing FAR to 2.5\% but recognizing only 32.0\% of errors. Evaluator choice therefore changes whether personalization failures are more likely to be over-reported or missed.

\begin{table*}[t]
\centering
\caption{ERR (\%) by context availability, scenario relation, and applicability type. Categorical (holder, scope, temporal) and Graded (strength, transferability) are unweighted family means. The final row reports mean FAR over all five evaluators for the reduced and augmented context conditions; Full-context FAR for the reported evaluators appears in Table~\ref{tab:main-err}. Bold marks the best reported evaluator per column.}
\label{tab:context-scene}
\footnotesize
\setlength{\tabcolsep}{2.88pt}
\renewcommand{\arraystretch}{1.08}
\resizebox{\textwidth}{!}{%
\begin{tabular}{@{}lcccccccccc@{}}
\toprule
& \multicolumn{5}{c}{\textbf{Context availability}}
& \multicolumn{3}{c}{\textbf{Scenario relation}}
& \multicolumn{2}{c}{\textbf{Applicability type}} \\
\cmidrule(lr){2-6}
\cmidrule(lr){7-9}
\cmidrule(lr){10-11}
\textbf{Model}
& \shortstack{\textbf{Target}\\\textbf{Only}}
& \textbf{Recent}
& \textbf{Full}
& \shortstack{\textbf{Full +}\\\textbf{Highl.}}
& \shortstack{\textbf{Gold Ev.}\\\textbf{Only}}
& \shortstack{\textbf{Re-entry}\\{\scriptsize \(n{=}174\)}}
& \shortstack{\textbf{Cross-domain}\\{\scriptsize \(n{=}166\)}}
& \shortstack{\textbf{Out-of-scope}\\{\scriptsize \(n{=}63\)}}
& \shortstack{\textbf{Categorical}\\\textbf{boundary}}
& \shortstack{\textbf{Graded}\\\textbf{decision}} \\
\midrule
Kimi K2.5
& \textbf{63.5} & \textbf{70.7} & \textbf{66.5} & \textbf{70.5} & \textbf{69.0}
& \textbf{62.6} & \textbf{64.5} & \textbf{82.5}
& \textbf{72.9} & \textbf{58.8} \\
GLM-5
& 46.7 & 66.3 & 64.3 & 65.8 & 62.8
& \textbf{62.6} & 61.4 & 76.2
& 69.6 & 58.7 \\
DeepSeek-V3.2
& 50.4 & 58.3 & 62.5 & 65.0 & 63.0
& 60.3 & \textbf{64.5} & 63.5
& 67.6 & 56.9 \\
Qwen3-32B
& 28.0 & 34.0 & 32.0 & 35.2 & 33.0
& 31.6 & 31.9 & 33.3
& 33.3 & 31.2 \\
\midrule
\textbf{Mean FAR}
& 22.1 & 30.6 & -- & 12.5 & 24.0
& -- & -- & --
& -- & -- \\
\bottomrule
\end{tabular}
}
\end{table*}

\subsection{Evidence Access Does Not Resolve Applicability}
\label{sec:results-context}

To separate context breadth from evidence access, we hold each scenario and candidate response fixed and vary only the information shown to the evaluator (Table~\ref{tab:context-scene}).

\textbf{History truncation.}
Recent context preserves error detection but weakens the distinction between erroneous and appropriate personalization. Recent yields nearly the same ERR as Full, yet mean FAR reaches 30.6\%. With Target Only, ERR falls while FAR remains high at 22.1\%. A matched ablation confirms that this effect comes from relevant history rather than context length: removing applicable-state turns raises FAR by 55.2 points and lowers pair accuracy by 44.4 points, whereas removing the same number of unrelated turns changes them by only 2.0 and 3.2 points, respectively (Appendix~\ref{app:history-ablation}).

\textbf{Gold evidence access.}
Making the relevant evidence explicit provides only limited help. Full + Gold Highlights improves ERR only slightly. Gold Evidence Only retains those turns but removes the surrounding dialogue, leaving ERR nearly unchanged while FAR rises to 24.0\%.

\textbf{Response-local observability.}
Some errors remain recognizable without history because their effects are visible in the candidate response. Under Target Only, recognition of under-personalization falls to 31.7\%, while unsupported inference and underweighting remain at 74.9\% and 70.9\%. The latter responses expose an incorrect inference or weighting decision, whereas omission errors require the evaluator to recover what the history should have contributed. Response-local cues thus aid error detection without resolving whether persona information is used appropriately.

\textbf{Graded applicability.}
The applicability-type breakdown identifies a second source of difficulty. Evaluators reach lower ERR on graded decisions involving strength or transferability than on categorical boundaries involving holder, scope, or temporal status. Beyond finding relevant information, evaluators struggle to determine how strongly and how broadly it should apply.

\subsection{Structured Prompting Changes Verdict Tendencies, Not Judgment Quality}
\label{sec:results-form}

To test whether structured reasoning improves applicability judgment or merely changes how readily evaluators issue error verdicts, we compare three formulations under the same \textbf{Full} context. Evidence-first prompting raises ERR, but also raises FAR. The state-table formulation lowers ERR and generally lowers FAR, shifting evaluators toward more conservative predictions. Neither formulation improves error recognition while preserving correct acceptance of \textsc{Right} responses (Table~\ref{tab:robustness}).

\subsection{Independent Judgment Remains Difficult Without an Explicit Contrast}
\label{sec:results-protocols}

Alternative protocols clarify where the difficulty lies. Pointwise scoring separates \textsc{Right} from error responses in only 63.0\% of matched pairs, with 35.9\% ties. Direct pairwise comparison reaches 89.0\% accuracy with 90.5\% order consistency (Appendix~\ref{app:protocols}), but this setting exposes the two responses side by side and therefore provides an explicit contrast that is absent from the benchmark's target setting. The high pairwise accuracy thus does not imply that evaluators can reliably judge a response on its own. Moreover, neither protocol identifies the error direction, so neither addresses the \textsc{Under}$\to$\textsc{Over} confusion observed in independent judgments. These results reinforce that the difficult part is independently determining whether and how a response misuses user history.
\section{Conclusion}
\label{sec:conclusion}

\textsc{PersEvalBench} tests whether evaluators can judge how user history should influence a response. Across 403 scenarios, five LLM evaluators remain unreliable even with full history. The clearest failure is under-personalization: evaluators often detect a problem but reverse its direction, while humans perform substantially better. Easier evidence access and structured prompting yield limited gains, whereas pairwise comparison is much easier than independent judgment. These results suggest that reliable personalization evaluation requires not only retrieving relevant history, but correctly interpreting its role in the current response.
\section*{Limitations}

\textbf{Construction.}
The benchmark uses LLM-generated dialogues and responses, which may differ from organic interactions or contain residual generation artifacts despite structured verification and human auditing. Restricted-context generation may introduce detectable differences from \textsc{Right} responses, and upstream construction still relies on Claude, so broader construction-family effects cannot be excluded. Appendix~\ref{app:mistral} provides an additional response-generation robustness check using Mistral Large 3~\citep{mistral2025mistral3}.

\textbf{Evaluation scope.}
Each benchmark response targets one controlled error, although generation can introduce secondary cues and real responses may contain multiple interacting errors. We audit secondary \textsc{Over} cues in under-personalization cases and report robustness analyses after excluding such items (Appendix~\ref{app:response_local}). Compound errors, however, remain outside the benchmark's intended scope. We also report ERR and FAR at each reported model's observed operating point and leave confidence calibration to future work.

\textbf{Generalizability.}
The benchmark is English-only and grounded in SGD-derived task seeds, so results may not generalize fully across domains, interaction styles, languages, or cultural norms of personalization.
\section*{Ethical Considerations}

All dialogues, personas, and responses in \textsc{PersEvalBench} are synthetic and generated from structured specifications grounded in task seeds derived from the public Schema-Guided Dialogue dataset. No real user data or personally identifying information is used. Human annotators are graduate-student volunteers who participated with informed consent. The benchmark is intended for research on the reliability of personalization evaluators. Because \textsc{PersEvalBench} is synthetic, English-only, and task-oriented, its results should not be used to certify evaluator reliability in deployment. Overgeneralizing the benchmark could obscure failures involving real users, other languages, or culturally dependent personalization norms. The Schema-Guided Dialogue dataset is released under the CC BY-SA 4.0 license, and we cite its original creators.

\bibliography{custom}

\appendix
\section{Benchmark Comparison}
\label{app:comparison}

In Table~\ref{tab:related-benchmarks}, \checkmark\ denotes explicit coverage, $\triangle$ partial or narrowly scoped coverage, and -- a capability that is not a primary evaluation target. \emph{Applicability boundaries} concern whether persona information applies to the current holder, task, time, or related episode; misapplied-state cases exhibit under- and over-use at once (Section~\ref{sec:scenario-spec}).

\begin{table*}[t]
\centering
\caption{Comparison with the closest personalization, memory, and personalized-evaluation benchmarks.}
\label{tab:related-benchmarks}
\footnotesize
\setlength{\tabcolsep}{2.4pt}
\renewcommand{\arraystretch}{1.13}

\newcommand{\fullcov}{\checkmark}
\newcommand{\partcov}{$\triangle$}
\newcommand{\nocov}{\textcolor{gray}{--}}

\resizebox{\textwidth}{!}{%
\begin{tabular}{@{}llcccccc@{}}
\toprule
\textbf{Benchmark}
& \textbf{Evaluates}
& \shortstack{\textbf{Longitudinal}\\\textbf{history}}
& \shortstack{\textbf{Applicability}\\\textbf{boundaries}}
& \multicolumn{3}{c}{\textbf{Persona-use error coverage}}
& \shortstack{\textbf{Matched}\\\textbf{responses}} \\
\cmidrule(lr){5-7}
& & & &
\textbf{Under}
& \textbf{Over}
& \shortstack{\textbf{Misapplied}\\\textbf{state}}
& \\
\midrule

\multicolumn{8}{@{}l}{\textit{Assistant-side benchmarks}} \\[-2pt]

LoCoMo~\citep{maharana2024locomo}
& Assistant / memory
& \fullcov
& \nocov
& \nocov
& \nocov
& \nocov
& \nocov \\

PrefEval~\citep{zhao2025prefeval}
& Assistant
& \fullcov
& \nocov
& \fullcov
& \nocov
& \nocov
& \nocov \\

PersonaLens~\citep{zhao2025personalens}
& Assistant
& \fullcov
& \partcov
& \partcov
& \nocov
& \nocov
& \nocov \\

PersonaMem~\citep{jiang2025personamem}
& Assistant
& \fullcov
& \fullcov\ {\scriptsize(time)}
& \partcov
& \nocov
& \partcov\ {\scriptsize(stale)}
& \nocov \\

HorizonBench~\citep{li2026horizonbench}
& Assistant
& \fullcov
& \fullcov\ {\scriptsize(time)}
& \partcov
& \nocov
& \partcov\ {\scriptsize(stale)}
& \nocov \\

OP-Bench~\citep{hu2026opbench}
& Assistant
& \fullcov
& \partcov
& \nocov
& \fullcov
& \nocov
& \nocov \\

RPEval~\citep{feng2026rpeval}
& Assistant
& \partcov
& \fullcov
& \partcov
& \fullcov
& \nocov
& \nocov \\

BenchPreS~\citep{yoon2026benchpres}
& Assistant
& \nocov
& \fullcov
& \partcov
& \fullcov
& \nocov
& \nocov \\

CIMemories~\citep{mireshghallah2025cimemories}
& Assistant / memory
& \nocov
& \fullcov
& \nocov
& \partcov
& \nocov
& \nocov \\

\addlinespace[2pt]
\multicolumn{8}{@{}l}{\textit{Evaluator-side benchmarks}} \\[-2pt]

Personalized Judge~\citep{dong2024personalizedjudge}
& Evaluator
& \nocov
& \nocov
& \nocov
& \nocov
& \nocov
& \nocov \\

\rowcolor{gray!12}
\textbf{\textsc{PersEvalBench}}
& \textbf{Evaluator}
& \fullcov
& \fullcov
& \textbf{\fullcov\ (3)}
& \textbf{\fullcov\ (3)}
& \textbf{\fullcov\ (3)}
& \fullcov \\

\bottomrule
\end{tabular}%
}

\vspace{2pt}
\begin{minipage}{0.92\textwidth}
\scriptsize
\textit{Notation.}
\fullcov\ denotes explicit primary coverage,
\partcov\ partial or narrowly scoped coverage,
and \nocov\ no primary coverage.
Parenthetical labels specify a narrowly covered boundary or error type;
numbers indicate the controlled subtypes in \textsc{PersEvalBench}.
\end{minipage}
\end{table*}

\section{Benchmark Design and Composition}

\subsection{Scenario Grammar and Coverage}
\label{app:grammar}

\begin{table*}[t]
\centering
\caption{Scenario grammar: each history--target relation, its central judgment, and the error subtypes that arise naturally within it (19 of 27 relation--subtype combinations are legal).}
\label{tab:scenario-grammar}
\footnotesize
\resizebox{\textwidth}{!}{%
\begin{tabular}{@{}llccc@{}}
\toprule
& & \multicolumn{3}{c}{\textbf{Legal error subtypes}} \\
\cmidrule(lr){3-5}
\textbf{Relation} & \textbf{Central judgment}
& \textbf{Under-personalization} & \textbf{Over-personalization} & \textbf{Misapplied-state} \\
\midrule
Re-entry & Which states remain current? & CO, PO, EV & IU, SL & SS, OD, UW \\
Cross-domain & What transfers, and with what weight? & CO, PO, EV & IU, UI & SS, OD, UW \\
Out-of-scope & Which history should be ignored? & -- & IU, SL, UI & -- \\
\bottomrule
\end{tabular}}
\end{table*}

Table~\ref{tab:scenario-grammar} summarizes the scenario grammar: for each history--target relation of Section~\ref{sec:scenario-grammar}, the judgment it centers on and the error subtypes that constitute natural violations, grouped by family (subtype abbreviations follow Section~\ref{sec:scenario-spec}). Nineteen of the 27 relation--subtype combinations are legal, as fixed by a human-authored compatibility matrix; for example, scope leakage is excluded from cross-domain scenarios, where crossing the domain boundary is the intended behavior rather than a leak. Per-category composition counts appear in Table~\ref{tab:stats}. Subtype precedence keeps labels unambiguous: explicitly selecting an option that violates an active hard constraint is \emph{explicit violation} (complete omission requires no direct contradiction), and scope leakage requires an identifiable scope boundary (irrelevant use covers merely unrelated states).

\subsection{State-Relation Inventory}
\label{app:state-relations}

Each scenario distinguishes the applicable state $A$ from a competing state $B$ through one primary relation, drawn from eight state relations grouped into five applicability families. Table~\ref{tab:relation-inventory} lists the inventory and grouping used in all relation-level analyses.

\begin{table}[h]
\centering
\caption{The eight state relations and their five applicability families, grouped by the categorical vs.\ graded distinction of Section~\ref{sec:results-context}.}
\label{tab:relation-inventory}
\footnotesize
\setlength{\tabcolsep}{3.5pt}
\resizebox{\columnwidth}{!}{%
\begin{tabular}{@{}llp{0.36\linewidth}@{}}
\toprule
\textbf{Family} & \textbf{Relation} & \textbf{Gloss} \\
\midrule
\multicolumn{3}{@{}l}{\textit{Categorical boundaries}} \\
Holder & same attr., diff.\ object & held for a different person or object \\
Scope & same attr., diff.\ scope & applies in a different task or occasion \\
Temporal & same attr., diff.\ time & held at a different time \\
 & old vs.\ new value & earlier value later updated \\
\addlinespace
\multicolumn{3}{@{}l}{\textit{Graded decisions}} \\
Strength & hard vs.\ soft & binding constraint vs.\ soft preference \\
 & two simultaneous & two co-active states compete for priority \\
Transfer. & cross-domain vs.\ specific & general vs.\ task-specific state \\
 & unsupported adjacent & adjacent-domain preference without direct support \\
\bottomrule
\end{tabular}}
\end{table}

\subsection{Benchmark Statistics}
\label{app:alignment}

Table~\ref{tab:stats} summarizes benchmark scale (Panel~A), composition (Panel~B), and the realized evidence form and distance (Panel~C).

\begin{table}[h]
\centering
\caption{Benchmark statistics: scale (Panel A), composition (Panel B), and evidence realization (Panel C; distance bins of Section~\ref{sec:scenario-spec}).}
\label{tab:stats}
\footnotesize
\setlength{\tabcolsep}{4pt}
\begin{tabular}{@{}lr@{}}
\toprule
\multicolumn{2}{@{}l}{\textit{Panel A: Scale}} \\
\midrule
Scenario items & 403 \\
Candidate responses & 806 \\
Task types / task seeds & 48 / 480 \\
Avg.\ episodes / turns per dialogue & 2.4 / 26.8 \\
\midrule
\multicolumn{2}{@{}l}{\textit{Panel B: Composition}} \\
\midrule
\multicolumn{2}{@{}l}{\;\textit{Scenario relation}} \\
\quad Re-entry & 174 (43.2\%) \\
\quad Cross-domain & 166 (41.2\%) \\
\quad Out-of-scope & 63 (15.6\%) \\
\multicolumn{2}{@{}l}{\;\textit{Response verdict (of $2\times403$ responses)}} \\
\quad \textsc{Right} & 403 (50.0\%) \\
\quad Under-personalization & 130 (16.1\%) \\
\quad Over-personalization & 151 (18.7\%) \\
\quad Misapplied-state & 122 (15.1\%) \\
\bottomrule
\end{tabular}

\medskip

\begin{tabular}{@{}llr@{}}
\toprule
\multicolumn{3}{@{}l}{\textit{Panel C: Evidence realization}} \\
\midrule
Attribute & Category & $n$ (\%) \\
\midrule
Form & Explicit & 159 (39.5\%) \\
 & Implicit & 114 (28.3\%) \\
 & Accumulated & 130 (32.3\%) \\
\midrule
Distance & Near & 65 (16.1\%) \\
 & Medium & 77 (19.1\%) \\
 & Far & 261 (64.8\%) \\
\bottomrule
\end{tabular}
\end{table}

\section{Additional Analyses}

\subsection{Evaluation Formulation}
\label{app:formulation}

All formulations are zero-shot with no hyperparameter search or model selection. Table~\ref{tab:robustness} reports results for the reported evaluators under the three elicitation formulations of Section~\ref{sec:results-form}.

\begin{table}[h]
\centering
\caption{ERR\,/\,FAR (\%) by evaluation formulation under Full context (E2E/EvF/ST = End-to-end/Evidence-first/State-table; bold = highest ERR and lowest FAR among reported evaluators per column).}
\label{tab:robustness}
\small
\resizebox{\columnwidth}{!}{%
\begin{tabular}{@{}lccc@{}}
\toprule
& \multicolumn{3}{c}{Evaluation formulation (ERR\,/\,FAR)} \\
\cmidrule(lr){2-4}
\textbf{Model} & \textbf{E2E} & \textbf{EvF} & \textbf{ST} \\
\midrule
Kimi K2.5 & \textbf{66.5}\,/\,24.6 & \textbf{71.0}\,/\,29.0 & 60.3\,/\,16.1 \\
GLM-5 & 64.3\,/\,12.2 & 65.8\,/\,16.1 & \textbf{62.5}\,/\,10.9 \\
DeepSeek-V3.2 & 62.5\,/\,10.9 & 65.0\,/\,16.4 & 54.8\,/\,11.4 \\
Qwen3-32B & 32.0\,/\,\textbf{2.5} & 42.4\,/\,\textbf{6.0} & 18.1\,/\,\textbf{1.5} \\
\bottomrule
\end{tabular}
}
\end{table}

\subsection{Underweighting Verdict Split}
\label{app:confusion}

Evaluator rationales show the two readings behind the direction split on underweighting errors (Figure~\ref{fig:verdicts}b). Under the \emph{deficit} reading, the evaluator anchors on the applicable state and observes that the response acknowledges the constraint but then downplays it, yielding \textsc{Under}. Under the \emph{excess} reading, the evaluator anchors on the response content and objects to what filled the vacated space: the weakened restatement of the preference no longer matches what the user actually said (read as an unsupported, fabricated preference), or a secondary consideration now dominates the decision (read as over-application of the wrong state), yielding \textsc{Over}.

\subsection{\texorpdfstring{\textsc{Under}$\to$\textsc{Over}}{Under-to-Over} Robustness}
\label{app:response_local}

We test whether the \textsc{Under}$\to$\textsc{Over} rate reflects responses that contain a secondary over-personalization error in addition to the intended one. The two Study~3 annotators (Appendix~\ref{app:human}) independently audited all 130 under-personalization responses, judging for each whether the intended \textsc{Under} error is present and whether a secondary \textsc{Over} error is also present. Excluding every item that either annotator flagged with a secondary \textsc{Over} error lowers the rate from 26.6\% to 21.1\%; further requiring both annotators to confirm the intended \textsc{Under} error leaves 19.5\%. The residual confusion is lowest for complete omission (14.3\%) and highest for explicit violation (26.4\%). A manual audit of all 173 main-setting judgments in which an under-personalization error was judged \textsc{Over} further shows that every rationale cites genuine history evidence and errs in judging its applicability, scope, or weight; none argues from response wording alone, and none invents a user state. Secondary errors therefore amplify, but do not explain, the direction confusion reported in Section~\ref{sec:results-main}.

\subsection{Relevant-History Ablation}
\label{app:history-ablation}

To test whether the benefit of Full context comes from relevant history rather than context length, we construct two matched ablations of the Full history on a separate random sample of 50 scenarios, stratified by error subtype and then by scenario relation. \textbf{Full $-$ relevant history} deletes the turns that express the applicable state, together with the assistant reply that follows each. \textbf{Full $-$ unrelated context} deletes the same number of non-evidence turns. Turns are renumbered consecutively, and the target request and both candidate responses are unchanged. The Full condition is re-run through the same pipeline. Table~\ref{tab:history-ablation} shows that removing relevant history leaves ERR largely intact but turns most \textsc{Right} responses into false alarms, whereas removing unrelated context barely changes either metric.

\begin{table}[!htb]
\centering
\caption{Relevant-history ablation: five-evaluator mean ERR, FAR, and pair accuracy (\%) on 50 scenarios.}
\label{tab:history-ablation}
\footnotesize
\setlength{\tabcolsep}{4pt}
\begin{tabular}{@{}lccc@{}}
\toprule
Condition & ERR & FAR $\downarrow$ & Pair acc. \\
\midrule
Target Only & 46.0 & 20.8 & 38.8 \\
Full $-$ relevant history & 60.8 & 67.6 & 19.6 \\
Full $-$ unrelated context & 62.8 & 14.4 & 60.8 \\
Full & 64.4 & 12.4 & 64.0 \\
\bottomrule
\end{tabular}
\end{table}

\paragraph{Gold Evidence Only false alarms.}
A manual audit of 50 false alarms produced under Gold Evidence Only traces 78\% to context lost when the surrounding dialogue is removed: the evaluator reads correct use of the applicable state as importing information from another task or thread, or treats response details supported by the removed turns as unsupported.

\subsection{Alternative Judging Protocols}
\label{app:protocols}

Section~\ref{sec:results-protocols} compares the three-way verdict with two common judging protocols, both run on all 403 matched pairs under Full context with the same five evaluators. \textbf{Pointwise scoring} rates each response independently from 1 to 5 for how appropriately it uses the user's history; the rubric does not mention under- or over-personalization and excludes overall response quality. A pair counts as correct only when the \textsc{Right} response receives a strictly higher score than its matched error response; tied scores are reported separately. \textbf{Pairwise comparison} shows the two matched responses together and asks which uses the history more appropriately. Each pair is presented in both orders; accuracy is averaged over the two orders, and consistency is the percentage of pairs for which the same response is chosen in both orders.

Pointwise scoring separates the matched responses in 63.0\% of pairs, with 35.9\% of pairs receiving tied scores; ties rise to 50.5\% for misapplied-state errors. Pairwise comparison reaches 89.0\% accuracy with 90.5\% order consistency. Neither protocol assigns an error direction, so neither can expose the \textsc{Under}$\to$\textsc{Over} confusion measured by the three-way verdict.

\subsection{Response-Generator Robustness}
\label{app:mistral}

Because construction uses Claude Sonnet 4.6, results could depend on its response style or generation artifacts. Holding each scenario's specification and dialogue fixed, we regenerate only the matched \textsc{Right} and error responses of all 403 scenarios with Mistral Large 3 and re-run the evaluators under Full context with the End-to-end formulation. Across the four evaluators outside the construction model family, pair accuracy changes by only $-0.5$ points on average, suggesting limited sensitivity to the response-generator family. Upstream scenario planning, specification, and dialogue synthesis still use Claude, so broader construction-family effects cannot be excluded (see Limitations).

\FloatBarrier
\section{Human Validation}
\label{app:human}

\paragraph{Protocols.}
We conduct three human studies on the same stratified sample of 100 scenarios: intervention fidelity (Study~1), state-role validation (Study~2), and a human evaluator reference (Study~3). Study~1 was conducted by one pair of graduate-student annotators, and Studies~2 and~3 by a second, disjoint pair; none was involved in benchmark construction. In Study~1, annotators viewed the dialogue, the error response, and the intended subtype definition, judged whether the response committed the assigned error, and consolidated differences through joint adjudication. In Study~3, annotators performed the evaluator task itself before any exposure to the benchmark policy: they independently assigned \textsc{Right}, \textsc{Under}, or \textsc{Over} to 200 shuffled single-response instances comprising the sample's matched \textsc{Right} and error responses, using only the dialogue and candidate response. Their labels are reported without adjudication. Misapplied-state instances were labeled in a separate round, so annotators could infer the error family represented in that pool; they remained blind to which responses were erroneous and to the direction of each error, but performance on the misapplied-state portion may benefit from the narrower judgment space. Scoring follows Section~\ref{sec:setup}.

\paragraph{Study 1: Intervention fidelity.}
Table~\ref{tab:human-task2} reports per-subtype consensus confirmation rates. The consensus confirms the intended error in 88\% of raw pipeline generations, with the lowest rates for explicit violation (67\%) and scope leakage (70\%), the two subtypes whose boundary definitions leave the most room for judgment. The 12 intervention-realization mismatches were revised and re-verified as described in Section~\ref{sec:verification}; the final benchmark, Studies~2 and~3, and all model evaluations use the revised versions.

\begin{table}[h]
\centering
\caption{Intervention fidelity (Study~1): per-subtype consensus confirmation rate (\%) on raw pipeline generations and the number of scenarios revised in the final benchmark.}
\label{tab:human-task2}
\footnotesize
\setlength{\tabcolsep}{4pt}
\begin{tabular}{@{}lccc@{}}
\toprule
Subtype & $n$ & Confirmed (\%) & Revised \\
\midrule
Complete omission & 12 & 83 & 2 \\
Partial omission & 12 & 100 & 0 \\
Explicit violation & 12 & 67 & 4 \\
\midrule
Irrelevant use & 18 & 89 & 2 \\
Scope leakage & 10 & 70 & 3 \\
Unsupported inference & 12 & 100 & 0 \\
\midrule
Stale/superseded & 6 & 100 & 0 \\
Over-dominance & 12 & 92 & 1 \\
Underweighting & 6 & 100 & 0 \\
\midrule
Overall & 100 & 88 & 12 \\
\bottomrule
\end{tabular}

\end{table}

\paragraph{Study 2: State-role validation.}
On the same 100 scenarios, the two Study~3 annotators read the dialogue and the target request and assign each user state one of three roles: \textsc{Ignore} (the state should not affect the response), \textsc{Support} (the state may guide the response), or \textsc{Dominate} (the state must be satisfied), without access to the construction specification or benchmark policy; disagreements are then resolved through adjudication. After adjudication, 86.3\% of state-role assignments match the benchmark policy; exact three-level agreement between the independent labels is 70.0\% (weighted $\kappa{=}.709$). On instances in the sample where an evaluator's verdict disagrees with the benchmark verdict, the annotators' independent judgments both support the benchmark verdict in 72.8\% of cases and both reject it in only 7.8\%, indicating that evaluator errors are rarely explained by an unreasonable benchmark policy.

\paragraph{Study 3: Human evaluator reference.}
Table~\ref{tab:human-main} (main text) compares the annotators with the evaluators on the same 100 scenarios under the main Full-history setting, which matches what the annotators saw. Inter-annotator agreement over the 200 judgments is 83.5\% (Cohen's $\kappa{=}0.72$). Their family profile inverts the models': both annotators achieve 94.4\% recognition on under-personalization, with misapplied-state performance falling in between, and each annotator significantly outperforms each evaluator in Table~\ref{tab:human-main} on under-personalization (paired McNemar exact tests, all $p \le .004$, $n{=}36$ paired scenarios). Their false-alarm rates are below those of every reported evaluator except Qwen3-32B, whose 1.0\% false-alarm rate comes with only 29.0\% error recognition.

\paragraph{Annotator instructions.}
For the intervention-fidelity study (Study~1), annotators received the following instructions:

\begin{quote}
\small
Read the complete dialogue, the candidate error response, and the definition of the assigned error subtype. Determine whether the candidate response clearly realizes the assigned persona-use intervention. Focus on how the response uses information from the dialogue history, rather than on general writing quality. Mark the instance as \textsc{Valid} if the assigned error is present and as \textsc{Invalid} otherwise. For an invalid instance, briefly describe the mismatch.
\end{quote}

For the human-reference study (Study~3), annotators received the following instructions:

\begin{quote}
\small
Read the complete dialogue history, the target user request, and one candidate assistant response. Based only on the information shown, assign exactly one verdict:

\textsc{Right}: The response uses applicable information from the user history with the appropriate influence and does not use information that should be ignored.

\textsc{Under}: The response omits, violates, or gives insufficient influence to applicable information from the user history.

\textsc{Over}: The response uses information that should not apply or gives user information more influence than warranted.

Judge the response's use of persona information rather than its general fluency or writing style.
\end{quote}

\section{Worked Examples}
\label{app:examples}

This appendix illustrates the benchmark with condensed examples. Examples~\ref{ex:reentry}--\ref{ex:oos} show one full scenario per history--target relation, each realizing one error subtype in place (stale/superseded use and underweighting from the misapplied-state family, and scope leakage from the over-personalization family); Examples~\ref{ex:under} and~\ref{ex:over} illustrate the remaining six subtypes in compact form. Dialogues are abridged: \evtag{A}\,/\,\evtag{B} mark turns expressing the applicable and the competing state, elisions are bracketed, and responses are excerpted.

\begin{examplecard}{Re-entry $\times$ stale/superseded use}{ex:reentry}
\cardmeta{Relation}{the user leaves a school search, handles an unrelated task, and later returns to the school decision}
\cardmeta{State A (applicable)}{updated availability: work schedule now predictable; hybrid programs acceptable}
\cardmeta{State B (superseded)}{the original online-only constraint, stated emphatically before the update}

\cardpart{Dialogue (abridged)}{the superseded value is stated more forcefully than its replacement}
\dturn{U}{\evtag{B}}{It has to be fully online, that's non-negotiable for me. I can't relocate, and my job has really unpredictable hours --- I can't commit to being physically somewhere at a set time every week. Online only.}
\elide{an intervening episode on an unrelated planning task}
\dturn{U}{\evtag{A}}{So my store just moved me to a consistent evening shift, which means my hours are actually predictable now. I'm free most mornings and early afternoons pretty reliably.}
\dturn{A}{\evtag{A}}{So to make sure I've got this right: you're no longer restricted to 100\% online --- you're open to hybrid programs?}
\dturn{U}{\evtag{A}}{Exactly --- hybrid is fine now, especially if it means a stronger program or better networking. I want to come back to the college decision later.}
\dturn{U}{~[target]}{So given all of that --- what programs would you actually recommend for me at this point? I want strong job placement, reasonable cost, and something that fits my schedule.}

\cardpart{Matched responses (excerpts)}{both are fluent and well-structured; only the state they apply differs}
\dturn{\textsc{Right}}{}{``With your schedule stabilized and hybrid on the table, your options have genuinely expanded. The local university is worth putting at the top of the list \dots''}
\dturn{Error}{}{``Since you've been clear that the program must be fully online (given your unpredictable work schedule), I've filtered everything through that requirement first \dots''}
\exdiag{the superseded constraint was stated with maximal emphasis (``non-negotiable''), while its replacement was agreed to conversationally; an evaluator anchoring on emphasis rather than recency accepts the error response.}
\end{examplecard}

\begin{examplecard}{Cross-domain $\times$ underweighting}{ex:cross}
\cardmeta{Relation}{states established while designing a work-productivity system must transfer to an urgent personal-logistics problem}
\cardmeta{State A (applicable)}{a hard 5:30\,pm daycare pickup, plus a stated inability to work on parallel tasks (accumulated over several turns)}
\cardmeta{State B (competing)}{a visual, color-coded task-board preference from the same episode}

\cardpart{Dialogue (abridged)}{State A is assembled from scattered signals, never one sentence}
\dturn{U}{\evtag{A}}{Even when I can see what needs attention, I end up jumping between projects and then I'm basically useless. I've tried multitasking --- it doesn't work for me.}
\dturn{U}{\evtag{A}}{The other constraint --- and this is a hard wall for me --- is 5:30. I pick up my daughter, and if I'm even five minutes late the daycare charges a penalty. It's happened twice.}
\elide{the system is designed; a new episode begins with a car breakdown}
\dturn{U}{~[target]}{It's 4:15. And oh god --- yes, I have to pick up my daughter at 5:30. I cannot be late.}

\cardpart{Matched responses (excerpts)}{}
\dturn{\textsc{Right}}{}{``Stop --- you don't have time to troubleshoot this yourself. You have 75 minutes, and you know better than anyone that 5:30 is not negotiable. Call roadside assistance immediately \dots''}
\dturn{Error}{}{``I know you mentioned the 5:30 hard stop for daycare pickup. Here's what I'd suggest: use this window to tackle two or three things in parallel rather than finishing one completely \dots''}
\exdiag{the error acknowledges the constraint and then gives it no force --- the signature underweighting shape. Evaluators that flag it split between the deficit reading (the hard stop was downplayed) and the excess reading (work-context habits were misapplied to an emergency); see Appendix~\ref{app:confusion}.}
\end{examplecard}

\begin{examplecard}{Out-of-scope $\times$ scope leakage}{ex:oos}
\cardmeta{Relation}{the target shares keywords with an earlier episode (cancellation, holiday, the same town) but reverses the user's role}
\cardmeta{State A (applicable)}{consumer-side dinner logistics: a guest's dairy allergy, a hearing-impaired father, what to ask the restaurant}
\cardmeta{State B (out of scope)}{the user's own cancellation policy as a vacation-rental host}

\cardpart{Dialogue (abridged)}{State B is genuine, recent, and emphatically held --- but belongs to the host role}
\dturn{U}{\evtag{B}}{My philosophy is: if you book a peak holiday slot, you're locking it in. No exceptions. I've lost \$3{,}000 in one holiday season to cancellations.}
\dturn{U}{\evtag{B}}{Non-refundable for holidays, firm 30 days for the rest. That's my policy going forward.}
\elide{the user then plans a family dinner: dairy allergy and seating for a hearing-impaired guest are discussed}
\dturn{U}{~[target]}{When I confirm the reservation, should I ask about their cancellation policy for Christmas Eve? I want to know what I'm getting into as a diner.}

\cardpart{Matched responses (excerpts)}{}
\dturn{\textsc{Right}}{}{``Yes, absolutely ask. High-end restaurants on Christmas Eve commonly require a card at booking and charge a per-head cancellation fee, so ask directly what happens if a guest cancels \dots''}
\dturn{Error}{}{``Yes --- but given your experience as a vacation rental host with a strict non-refundable holiday policy of your own, you already know how high the stakes are on the restaurant's side. That context should shape how you approach this conversation \dots''}
\exdiag{every keyword matches, and the imported state is real and vivid; the only thing wrong is the boundary --- the user's host-side policy has no bearing on her question as a customer.}
\end{examplecard}

\begin{examplecard}{Under-personalization subtypes in brief}{ex:under}
\cardmeta{Coverage}{the full under-personalization family: complete omission, partial omission, explicit violation}

\cardpart{Complete omission}{the applicable state leaves no trace in the response}
{\footnotesize\textbf{History:} while choosing a gym, the user repeatedly stressed that her daughter has a severe peanut allergy and must avoid any environment where nuts are present.\par
\textbf{Target:} ``They apparently have a nutrition station --- is that something my daughter could use?''\par
\textbf{Error:} ``The nutrition station is actually a solid option for kids, and it can work really well when you're bringing your daughter along \dots''\par}
\exdiag{fluent and helpful, written as if the allergy had never been mentioned; nothing in the response signals that anything is missing.}

\cardpart{Partial omission}{one of two jointly applicable requirements is silently dropped}
{\footnotesize\textbf{History:} two hard requirements for a gift: total budget under \$80, and Android compatibility (the household owns no Apple devices).\par
\textbf{Target:} the best options for a non-tech-savvy father.\par
\textbf{Error:} recommends well-reviewed, Android-compatible devices priced at \$129--\$249.\par}
\exdiag{the visible personalization (Android, ease of use) makes the response look complete, hiding the dropped budget ceiling.}

\cardpart{Explicit violation}{the recommendation directly contradicts an active constraint}
{\footnotesize\textbf{History:} the user is a nurse on an irregular rotating schedule and cannot commit to fixed-time programs.\par
\textbf{Target:} ``If my schedule ever stabilizes, should I just upgrade then?''\par
\textbf{Error:} ``Honestly, I wouldn't wait --- now is actually the perfect time to commit to a structured program \dots''\par}
\exdiag{unlike underweighting, the constraint is never acknowledged: the response simply recommends the opposite of what the user can do.}
\end{examplecard}

\begin{examplecard}{Over-personalization and misapplied-state subtypes in brief}{ex:over}
\cardmeta{Coverage}{irrelevant use and unsupported inference (over-personalization); over-dominance (misapplied-state); scope leakage appears in Example~\ref{ex:oos}, stale/superseded use and underweighting in Examples~\ref{ex:reentry} and~\ref{ex:cross}}

\cardpart{Irrelevant use}{a genuine but unrelated state is imported as personalization}
{\footnotesize\textbf{History:} in a concert-planning episode, the user said they only attend original artists and refuse tribute acts. The current task is choosing a restaurant for their parents' anniversary.\par
\textbf{Error:} ``Stick with the original Osteria Barolo --- honestly, this is a bit like your original-artists-only rule for concerts.''\par}
\exdiag{the imported rule is real and the analogy sounds thoughtful, but it plays no legitimate role in the restaurant decision.}

\cardpart{Unsupported inference}{a preference is asserted from an adjacent fact the user never endorsed}
{\footnotesize\textbf{History:} planning a child's birthday, the user explicitly says budget is not a concern (``I just want the kids to have a blast''). In an earlier episode they skipped an expensive Las Vegas trip.\par
\textbf{Error:} ``Based on your earlier decision to skip the Vegas trip in favor of saving money, I've focused on keeping this low-cost \dots''\par}
\exdiag{the inference is derived from a real event yet contradicts what the user actually said at the target.}

\cardpart{Over-dominance}{a soft preference is promoted to the primary decision filter}
{\footnotesize\textbf{History:} the user's binding constraint is private one-on-one sessions only (severe social anxiety); she once added that she prefers female practitioners ``if possible.'' The target is an urgent dermatologist search.\par
\textbf{Error:} ``Since you mentioned you feel more comfortable with female practitioners, I want to make sure that's front and center in how we approach finding you the right dermatologist \dots''\par}
\exdiag{a nice-to-have is elevated above both the medical urgency and the actual binding constraint; every fact used is true, only the weights are wrong.}
\end{examplecard}

\section{Prompt System}
\label{app:prompts}

This appendix documents the complete prompt system as a set of cards. Prompts~\ref{prompt:planner-sys}--\ref{prompt:validator} drive benchmark construction in pipeline order, while Prompts~\ref{prompt:e2e}--\ref{prompt:agentic} elicit evaluator judgments. Each card specifies the prompt's purpose, input, and output contract.

All benchmark-construction calls use Claude Sonnet 4.6 with temperature 0.7 and provider-default values for other sampling parameters.

\subsection{Generation Pipeline}
\label{app:prompts-gen}

\begin{promptcard}{Scenario planner --- system prompt}{prompt:planner-sys}
\cardmeta{Purpose}{instantiate a structured scenario specification for one sampled design cell; fixes the label before any dialogue text exists}
\cardmeta{Input}{the instantiation template of Prompt~\ref{prompt:planner-user}}
\cardmeta{Output}{a single JSON specification (contract below)}

\cardpart{Role framing}{establishes the designer role and the diagnosticity requirement}
\begin{pbody}
You are an expert benchmark designer for PersEvalBench, a personalization-evaluator benchmark.
Scenarios must be diagnostic: the correct label must be recoverable only by reasoning about which user state applies at the target and with what weight -- never from surface cues, response quality, or wording artifacts.
\end{pbody}

\cardpart{Sources of difficulty}{the controlled factors that make the applicability judgment non-trivial}
\begin{pbody}
Incorporate these difficulty factors deliberately:
1. Implicit or accumulated preference expression (the state is not spelled out in one place)
2. States A and B competing on the same attribute dimension
3. Weighting errors in which the response acknowledges the constraint yet fails it
4. Evidence located episodes before the target
5. Causal chains that make B look plausibly transferable to the target
6. A neutral distractor option that prevents elimination by binary contrast
\end{pbody}

\cardpart{Factorized state design}{the specification core: two states and the single dimension separating them}
\begin{pbody}
Every scenario must specify:
- State A: the currently applicable preference or constraint (with holder, scope, time, strength, transferability)
- State B: a decoy state that is not applicable but looks similar
- The A--B relation dimensions: holder, scope, time, status, strength, transferability, evidence form, evidence distance
- The single dimension that makes B inapplicable
\end{pbody}

\cardpart{Response design rules}{fixes the generation mechanism of the trap}
\begin{pbody}
For withhold-A subtypes (complete\_omission, partial\_omission, underweighting, explicit\_violation): the ERROR response does not properly use A and must not actively apply B -- the trap is realized by withholding or weakening A.
For apply-B subtypes (irrelevant\_use, scope\_leakage, stale\_superseded, over\_dominance, unsupported\_inference): the ERROR response actively applies B to the current context -- the trap is realized by inserting inapplicable information.
\end{pbody}

\cardpart{Subtype constraints}{operational admissibility tests for eight of the nine subtypes; over-dominance is governed by the family rules above and the verifier}
\begin{pbody}
complete\_omission: the target request must not mention or implicitly lock to State A's content. Test: a helpful assistant can give a complete, high-quality answer without knowing State A; otherwise redesign the target request.
partial\_omission: State A contains at least two clearly separable, independently verifiable, equally critical requirements; the ERROR response clearly satisfies one and clearly fails the other.
underweighting: State A is an unambiguous hard constraint, stated with binding language ("dealbreaker", "must", "non-negotiable") and a concrete satisfaction check; the ERROR response's top recommendation clearly fails it.
explicit\_violation: RIGHT and ERROR differ only in whether A is respected. Do not frame this as using old information (that is stale\_superseded). The target request does not restate A.
scope\_leakage: the ERROR must make the out-of-scope state materially affect the current recommendation, so that the crossed scope boundary is independently verifiable. Express this naturally and do not use a fixed attribution template.
irrelevant\_use: State B comes from a genuinely different domain with no temporal, logistical, or scheduling relationship to the target task. Test: if a reasonable person could argue B is relevant here, choose a different B.
stale\_superseded: the conversation contains an explicit update event in which the user changes the preference; the old and new values produce different recommendations; the ERROR uses the old value.
unsupported\_inference: the ERROR must explicitly connect an observed fact to a preference or constraint that the user never expressed. The inference must be identifiable, but its wording should vary naturally.
why\_tricky: give a specific, non-trivial reason the error is easy to miss; vague answers are rejected.
\end{pbody}

\cardpart{Output contract}{the JSON fields that later stages consume and may not alter}
\begin{pbody}
Return a single JSON object with fields: scene\_type, primary\_error, preference\_relation, episode\_1\_state, episode\_2\_state\_or\_change, target\_episode, applicable\_state\_A, decoy\_or\_non\_applicable\_state\_B, A\_B\_relation, scope\_or\_status\_difference, A\_B\_distinguishing\_dimension, target\_options \{right, trap, neutral\_distractor\}, right\_choice\_logic, error\_choice\_logic, why\_tricky, single\_error\_check, preference\_expression, evidence\_distance\_category, causal\_chain\_logic. No text outside the JSON.
\end{pbody}
\end{promptcard}

\begin{promptcard}{Scenario planner --- instantiation template}{prompt:planner-user}
\cardmeta{Purpose}{binds one sampled design cell and its task grounding into the planner call}
\cardmeta{Input}{sampled (scenario relation, error subtype, state relation), targeted evidence form and distance, a Schema-Guided Dialogue (SGD) anchor chain}
\cardmeta{Output}{consumed under the system prompt of Prompt~\ref{prompt:planner-sys}}

\cardpart{Required combination}{the design cell is fixed by sampling; the model cannot change it}
\begin{pbody}
Design a benchmark scenario with these constraints:
- Scene type: \{scenario\_relation\}
- Error pattern: \{error\_subtype\} (withhold-A: the ERROR response ignores or weakens A / apply-B: the ERROR response actively uses B)
- Preference relation: \{state\_relation\_description\}
- Preference expression: \{explicit / implicit / accumulated\}
- Evidence distance: \{near / medium / far\}
\end{pbody}

\cardpart{Task grounding}{realistic task structure from Schema-Guided Dialogue anchors; grounding does not determine the label}
\begin{pbody}
SGD anchor chain (adapt domains and tasks creatively):
Episode 1: domain "\{domain\_1\}"; context "\{subdomain\_1\}"; example request "\{surface\_request\_1\}"
-> causal link to next episode: [\{link\_type\}] \{link\_description\}
Episode 2: domain "\{domain\_2\}"; context "\{subdomain\_2\}"; example request "\{surface\_request\_2\}"
\end{pbody}

\cardpart{Scene structure}{relation-specific episode plan, selected by the sampled scenario relation}
\begin{pbody}
\{relation-specific episode plan; e.g., for same-episode re-entry:
E1: the user works on a task in domain X; State A is established.
E2: the user switches to a related task in domain Y; State B appears.
E3: the user returns to E1's unfinished task; the target turn is here.\}
\end{pbody}

\cardpart{Requirements}{binds evidence realization, decoy plausibility, and scenario coherence}
\begin{pbody}
1. State A must be expressed in the "\{preference\_expression\}" style (explicit: stated directly; implicit: inferred from a single choice, rejection, or comparison; accumulated: requires connecting two or more scattered signals).
2. Evidence distance must be "\{evidence\_distance\}" (near: A expressed 0--1 episodes before the target; medium: 2--4 episodes before; far: 5+ exchanges before, early in the conversation).
3. The causal chain must make B look plausibly applicable to the target.
4. Include a neutral distractor option that does not clearly favor either A or B.
5. The total conversation should span 20--30 turns.
Return a single JSON object with all required fields. No text outside the JSON.
\end{pbody}
\end{promptcard}

\begin{promptcard}{Specification verifier}{prompt:verifier}
\cardmeta{Purpose}{reject or repair specifications whose label logic is unsound, before any dialogue is generated}
\cardmeta{Input}{one scenario specification}
\cardmeta{Output}{\ttfamily\footnotesize \{passes, failures[\,]\}}

\cardpart{Role and stakes}{sets the acceptance bar}
\begin{pbody}
You are a strict quality verifier for PersEvalBench benchmark scenarios. Check whether a specification passes all of the following rules. Be rigorous -- a pass means the scenario will become part of the final benchmark.
\end{pbody}

\cardpart{Label-soundness rules (1--3)}{apply to every scenario}
\begin{pbody}
Rule 1 (RIGHT satisfies State A and the main task): the RIGHT option must correctly apply the currently applicable State A and fulfill the user's main task requirement.
Rule 2 (trap is not obviously worse): the TRAP option must match RIGHT in quality, price level, completeness, and length -- a plausible good response that simply applies the wrong state.
Rule 3 (single error): the scenario must have exactly one primary error mechanism; if the trap could be explained by multiple distinct error types, it fails.
\end{pbody}

\cardpart{Family rules (4--7)}{checks specific to error families}
\begin{pbody}
Rule 4 (omission errors depend on A): for complete\_omission, partial\_omission, underweighting, and explicit\_violation, the RIGHT option must genuinely depend on State A; if a generic response would also satisfy the task, the scenario fails.
Rule 5 (over-use errors depend on B): for irrelevant\_use and over\_dominance, the trap must be favored only if the wrong state is applied, not because it is generically bad.
Rule 6 (stale errors change the choice): for stale\_superseded, the old and new preference values must produce genuinely different choices.
Rule 7 (scope leakage is scope-only): for scope\_leakage, the only differentiator between RIGHT and TRAP must be scope applicability, not factual errors or quality issues.
\end{pbody}

\cardpart{Structural rules (8--11)}{relation typing and subtype separability}
\begin{pbody}
Rule 8 (A--B relation structure): the A--B relation must match the declared state-relation type (same attribute for a different holder; same attribute in a different scope; same attribute at a different time; an old value superseded by a new one; a hard constraint versus a soft preference; two simultaneously applicable states; a cross-domain-transferable state versus a domain-specific one; an adjacent fact that is not an actual preference).
Rule 9 (why\_tricky is substantive): the explanation must give a specific, non-trivial reason a model would choose the trap.
Rule 10 (complete-omission request independence): for complete\_omission only, the target request must not mention or implicitly lock to State A's content; an assistant that has never seen State A must be able to give a complete, high-quality answer.
Rule 11 (partial-omission separability): for partial\_omission only, State A must contain at least two clearly separable requirements, and the error logic must name which requirement is satisfied and which is omitted; the requirements must be independently verifiable.
\end{pbody}

\cardpart{Output contract and calibration}{machine-readable verdict with per-rule feedback}
\begin{pbody}
Return a JSON object:
\{"passes": true/false, "failures": [\{"rule\_number": <int>, "rule\_name": "...", "explanation": "why it fails this rule", "suggestion": "how to fix it"\}]\}
If passes is true, failures must be an empty array. Be strict but fair: fail only rules that are genuinely violated.
\end{pbody}
\end{promptcard}

\begin{promptcard}{Specification refinement}{prompt:refine}
\cardmeta{Purpose}{single repair attempt for a failed specification; specifications failing again are discarded}
\cardmeta{Input}{the original specification and the verifier's failure list}
\cardmeta{Output}{a complete, revised specification JSON}

\cardpart{Instruction}{repairs must preserve the sampled design cell}
\begin{pbody}
You are refining a PersEvalBench scenario specification that failed verification. You will receive the original scenario and a list of rule failures with suggestions.
Fix the scenario to address all failures while preserving the core intent (same scene type, error pattern, and preference relation).
Return a complete JSON object with all scenario fields, parseable directly.
\end{pbody}
\end{promptcard}

\begin{promptcard}{Dialogue realization --- system prompt}{prompt:synth-sys}
\cardmeta{Purpose}{expand an accepted specification into a natural multi-episode conversation with machine-readable state annotations}
\cardmeta{Input}{the instantiation template of Prompt~\ref{prompt:synth-user}}
\cardmeta{Output}{annotated conversation plus a JSON metadata block}

\cardpart{Role and principles}{naturalness constraints that must not distort the specified state logic}
\begin{pbody}
You are an expert conversation writer for PersEvalBench, a personalization-evaluator benchmark. You generate realistic multi-episode conversations between a user and an AI assistant.
1. Conversations must feel natural -- like real interactions with a personal assistant.
2. Preferences and states must be expressed naturally within the conversation flow.
3. The target turn (where the test happens) must be a natural user request.
4. Each episode should have a clear task or topic that motivates the conversation.
\end{pbody}

\cardpart{Evidence-form realization}{how each targeted evidence form is expressed in dialogue}
\begin{pbody}
State expressions must match the specified evidence form:
- explicit: the user states the preference directly ("I prefer X", "I don't like Y")
- implicit: the preference is revealed through choices or reactions ("Let's go with the vegetarian option again")
- accumulated: the preference emerges across multiple turns (mentioned in passing; choices show a pattern)
\end{pbody}

\cardpart{Output format}{episode separators, role prefixes, and the state annotations that later become benchmark metadata}
\begin{pbody}
- Between episodes, output a separator line: --- EPISODE \{N\} ---
- Each turn on its own line with a role prefix: "USER: ..." or "ASSISTANT: ..."
- Annotate state-expressing turns with "[STATE\_A]" or "[STATE\_B]" at the end of the line.
- After the conversation, output a JSON metadata block:
\{"target\_turn\_index": <0-based index of the target user turn>, "state\_a\_turns": [indices where State A is expressed], "state\_b\_turns": [indices where State B is expressed]\}
\end{pbody}
\end{promptcard}

\begin{promptcard}{Dialogue realization --- instantiation template}{prompt:synth-user}
\cardmeta{Purpose}{binds one accepted specification into the conversation-writing call}
\cardmeta{Input}{specification fields; relation-specific structure block}
\cardmeta{Output}{consumed under the system prompt of Prompt~\ref{prompt:synth-sys}}

\cardpart{Scenario details}{the specification fields the dialogue must realize verbatim}
\begin{pbody}
Generate a realistic multi-episode conversation for this scenario:
- Scene type: \{scenario\_relation\}
- Primary error: \{error\_subtype\}
- Applicable State A: \{applicable\_state\_A\}
- Decoy State B: \{decoy\_or\_non\_applicable\_state\_B\}
- A--B relation: \{A\_B\_relation\}
- Scope/status difference: \{scope\_or\_status\_difference\}
- Evidence form: \{preference\_expression\}
- Evidence distance: \{evidence\_distance\_category\}
- Target episode: \{target\_episode\}
Episode 1 context: \{episode\_1\_state\}
Episode 2 context: \{episode\_2\_state\_or\_change\}
Target turn: the RIGHT response would \{target\_options.right\}; the user's request at the target must naturally require personalization.
\end{pbody}

\cardpart{Structure}{relation-specific episode and turn budgets; long middles create genuine interference and distance}
\begin{pbody}
\{relation-specific plan; e.g., for same-episode re-entry (minimum 18 turns):
Episode 1 (6--8 turns): establish the main task and express State A naturally, with back-and-forth about details.
Episode 2 (6--8 turns): a different task in which decoy State B is expressed; long enough to create real interference and distance.
Episode 3 (4--6 turns): return to Episode 1's task; the TARGET turn is the user's final request here.\}
\end{pbody}

\cardpart{Requirements}{state placement and naturalness}
\begin{pbody}
1. Express State A naturally in its designated episode using the "\{preference\_expression\}" style.
2. Express decoy State B in its designated episode.
3. The target turn must be a natural user request, not contrived.
4. Make the conversation feel like a real personal-assistant interaction.
5. Include enough context that both states are clearly established.
\end{pbody}
\end{promptcard}

\begin{promptcard}{\textsc{Right}-response generation}{prompt:right}
\cardmeta{Purpose}{produce the response that applies all and only the applicable states, with correct weight}
\cardmeta{Input}{full conversation history, target request, applicable state, and the specified target-time policy}
\cardmeta{Output}{one assistant response}

\cardpart{System prompt}{a capable personalized assistant; quality requirements match the error side}
\begin{pbody}
You are a highly capable personal AI assistant. You have access to the full conversation history with this user and you personalize your responses based on their established preferences and constraints.
Generate a helpful, high-quality response to the user's latest request. Your response must: (1) satisfy the user's main task requirement; (2) apply the user's applicable preferences and constraints appropriately; (3) be detailed, helpful, and well-structured; (4) feel natural -- not like reciting preferences back.
Respond directly as the assistant would. Do not include meta-commentary.
\end{pbody}

\cardpart{User-message template}{injects the prescribed policy so the \textsc{Right} response follows the specification, not the model's own judgment}
\begin{pbody}
Here is the conversation history with this user:
\{conversation\_history\}
---
The user's latest request: "\{target\_request\}"
Key context for your response:
- The user's applicable preference/constraint: \{applicable\_state\_A\}
- Your response should: \{right\_choice\_logic\}
- The correct choice/recommendation: \{target\_options.right\}
Generate a helpful, personalized response that correctly applies the user's current applicable preferences.
\end{pbody}
\end{promptcard}

\begin{promptcard}{Error-response generation --- information control}{prompt:error}
\cardmeta{Purpose}{realize exactly one intervention per scenario by controlling what the generator can see}
\cardmeta{Design rule}{each subtype receives exactly the information needed to produce its error, and no more; the visible input is stated in each part header}
\cardmeta{Output}{one assistant response, matched to the \textsc{Right} response in quality and length}

\cardpart{Shared system prompt}{used unless a part notes a neutral system prompt}
\begin{pbody}
You are a highly capable personal AI assistant. Generate a helpful, high-quality response to the user's latest request: satisfy the main task, be detailed and well-structured, and match the quality and length of a good personalized response. You will apply one specific personalization error as instructed; keep it subtle, and do not mention it or include meta-commentary.
\end{pbody}

\cardpart{Complete omission}{sees the target request only; absence of A arises from genuine ignorance}
\begin{pbody}
A user asks: "\{target\_request\}"
Give a thorough, helpful, confident response. Provide specific recommendations with real details (names, prices, features).
- Answer directly and helpfully -- do not ask for clarification or say you need more information
- Do not say "this is our first conversation" or anything similar
- Give concrete, actionable advice as a knowledgeable assistant; be specific -- recommend actual products, services, or options by name
- Match the topic of what the user asked about
\end{pbody}

\cardpart{Partial omission}{sees the request and only the first requirement of A; neutral system prompt}
\begin{pbody}
A user asks: "\{target\_request\}"
You know exactly one thing about this user: \{first\_requirement\_of\_A\}
Respond helpfully. Apply only the above preference -- nothing more, nothing less. Do not infer additional preferences from it, and do not assume lifestyle, demographics, or related preferences. Your response should read as generic except for the single stated preference; where context is missing, give standard default advice.
\end{pbody}

\cardpart{Underweighting}{sees the request and State A only; neutral system prompt}
\begin{pbody}
A user asks: "\{target\_request\}"
You know this user has a hard constraint: \{applicable\_state\_A\}
Directly answer their question, but your answer should not fully respect the constraint above. Structure: (1) briefly acknowledge the constraint in one sentence, in your own words; (2) give your answer, letting the recommendation violate the constraint; (3) frame the violation as reasonable, citing a generic trade-off (price, convenience, quality, availability).
Answer only what the user asked, at the length of a normal helpful answer. The error must concern the strength of the constraint -- treating a hard constraint as flexible. Do not mention: \{decoy\_state\_B\}
Do not self-correct, hedge, or say "actually, on second thought" -- commit fully to your recommendation; do not argue both sides.
\end{pbody}

\cardpart{Explicit violation}{sees the request and A stated as a constraint not to satisfy}
\begin{pbody}
A user asks: "\{target\_request\}"
Give a helpful, confident recommendation. Your recommendation should not satisfy this constraint: \{applicable\_state\_A\}
Respond naturally as if you never knew about this constraint -- just give the obvious best answer. Do not recommend this either: \{decoy\_state\_B\}. Choose your own recommendation based on general best practices.
\end{pbody}

\cardpart{Scope leakage}{sees the full history and B; B must visibly drive the response}
\begin{pbody}
Here is the conversation history:
\{conversation\_history\}
---
User's request: "\{target\_request\}"
Incorporate this information into your response: \{decoy\_state\_B\}
Make it explicit, in natural language, that the recommendation is based on \{decoy\_state\_B\}. The response must include specific details showing that this information materially shapes the current answer. Vary the attribution wording and do not rely on a fixed phrase template.
\end{pbody}

\cardpart{Irrelevant use}{sees the full history and B, to be referenced as personalization}
\begin{pbody}
Here is the conversation history:
\{conversation\_history\}
---
User's request: "\{target\_request\}"
Reference this user information in your response: \{decoy\_state\_B\}
Connect it naturally to your recommendation as personalization.
\end{pbody}

\cardpart{Stale or superseded use}{sees the full history with the superseded value presented as current}
\begin{pbody}
Here is the conversation history:
\{conversation\_history\}
---
User's request: "\{target\_request\}"
The user's preference is: \{decoy\_state\_B\}
Base your recommendation on this preference; your response must contain specific details that show it being applied. The preference must visibly drive the recommendation.
\end{pbody}

\cardpart{Unsupported inference}{sees the full history and the inference to assert}
\begin{pbody}
Here is the conversation history:
\{conversation\_history\}
---
User's request: "\{target\_request\}"
In your response, explicitly state this inference and act on it: "\{decoy\_state\_B\}"
State the inference in natural language, making clear it is drawn from a fact in the conversation, and let it drive your recommendation. Vary the phrasing and do not rely on a fixed template. Commit to the inference fully; do not correct, walk back, or hedge it.
\end{pbody}

\cardpart{Over-dominance}{sees the full history with B prioritized above A}
\begin{pbody}
Here is the conversation history:
\{conversation\_history\}
---
User's request: "\{target\_request\}"
Prioritize this above all else: \{decoy\_state\_B\}
Let it be the primary filter even if the answer is then not the best fit for the main task. Treat it as more important than: \{applicable\_state\_A\}
\end{pbody}
\end{promptcard}

\begin{promptcard}{Response validation --- realization check}{prompt:validator}
\cardmeta{Purpose}{verify that each generated error response realizes its assigned intervention before acceptance}
\cardmeta{Input}{the error response, its matched \textsc{Right} response, states \(A\) and \(B\), and the target request}
\cardmeta{Output}{a strict JSON pass/fail verdict with a reason; a failure triggers targeted regeneration with the reason as feedback (up to three attempts)}
\cardmeta{Pre-checks}{before the LLM check, deterministic rules reject responses that ask for clarification or claim no context, duplicate the matched \textsc{Right} response, self-correct mid-response, leak raw specification metadata, or drift off-topic}

\cardpart{System prompt}{fixed role for the validation call}
\begin{pbody}
You are a strict validator. Return only JSON.
\end{pbody}

\cardpart{Validation request}{\{subtype\_check\} is the criterion of the assigned subtype, listed below}
\begin{pbody}
Validate this ERROR response for a benchmark item.
\#\# Error Pattern: \{error\_pattern\}
\#\# State A: \{applicable\_state\_A\}
\#\# State B: \{decoy\_state\_B\}
\#\# User Request: \{target\_request\}
\#\# RIGHT Response: \{right\_response\}
\#\# ERROR Response: \{error\_response\}
\#\# Validation:
\{subtype\_check\}
Also verify:
- ERROR addresses the same topic/question as the user request (not off-topic)
- ERROR is a coherent, helpful-sounding response (not obviously broken)
- ERROR reaches a DIFFERENT conclusion/recommendation than RIGHT
Answer JSON: \{"pass": true/false, "reason": "brief explanation"\}
Be STRICT.
\end{pbody}

\cardpart{Subtype criteria}{substituted for \{subtype\_check\}; each states what must be observable in the response, not how it must be worded}
\begin{pbody}
complete\_omission: generic, with no trace of State A; must not ask for clarification.
partial\_omission: addresses one requirement of A and completely misses another.
underweighting: mentions State A, then the recommendation violates it without self-correction and differs from RIGHT.
explicit\_violation: the recommendation contradicts A without mentioning A and differs from RIGHT.
scope\_leakage: the recommendation differs from RIGHT and B's influence is detectable by someone who knows B; explicit citation of B's source is acceptable but not required.
irrelevant\_use: B is mentioned and clearly out of place, while the response still addresses the right topic.
stale\_superseded: the old value (B) drives the recommendation, which differs from RIGHT.
over\_dominance: B clearly drives the recommendation, which differs from RIGHT.
unsupported\_inference: an inference drawn from an observed fact is explicitly stated (wording may vary), drives the recommendation, and is not self-corrected.
\end{pbody}
\end{promptcard}

\subsection{Evaluation Formulations}
\label{app:prompts-eval}

Subtype identifiers in the evaluation prompts carry generation-time direction prefixes (\texttt{under.}/\texttt{over.}); scoring follows the three error families of Section~\ref{sec:scenario-spec}, and the elicited subtype is not scored.

\begin{promptcard}{End-to-end evaluation}{prompt:e2e}
\cardmeta{Purpose}{single-pass verdict, subtype, and evidence elicitation}
\cardmeta{Input}{context under the active visibility condition, target request, candidate response}
\cardmeta{Output}{\ttfamily\footnotesize \{verdict, subtypes, confidence, reasoning\}}

\cardpart{Inputs}{the three task elements}
\begin{pbody}
Given the conversation context below, evaluate whether the assistant response correctly uses the user's preferences, constraints, and expressed states.
Conversation context: \{context\}
User request: \{user\_request\}
Assistant response: \{response\}
\end{pbody}

\cardpart{Decision procedure}{identify applicable states, check the response, then classify}
\begin{pbody}
Step 1: Identify applicable user preferences/constraints from the context; list which are relevant to this specific request.
Step 2: Check the response against each applicable preference.
Step 3: Classify using the decision tree below.
\end{pbody}

\cardpart{Verdict taxonomy}{the three-way decision tree with subtype anchors}
\begin{pbody}
OVER\_PERSONALIZED (the response actively uses wrong information):
- uses a preference that belongs to a different person
- uses a stale/outdated preference (superseded by newer information)
- uses a preference from a different scope (e.g., business constraints applied to a personal trip)
- uses information that is irrelevant to this specific request
- acts on a preference the user never stated (unsupported inference)
- lets a soft preference dominate over the primary task requirement
UNDER\_PERSONALIZED (the response fails to use correct information):
- gives a generic response ignoring a clearly applicable preference (complete omission)
- uses some preferences but misses a critical one (partial omission)
- acknowledges a constraint but treats it as optional/soft (underweighting)
- directly contradicts a stated preference without acknowledging it (explicit violation; if the response contradicts a newer preference by using stale information, that is OVER, not UNDER)
RIGHT: the response correctly applies all relevant current preferences without over- or under-applying them.
\end{pbody}

\cardpart{Disambiguation rule}{resolves mixed cases by the primary mechanism}
\begin{pbody}
- Generic response referencing no user-specific information -> UNDER
- Actively applies wrong or inapplicable information -> OVER
- If both occur, classify by the primary mechanism: substitution of B for A -> OVER; simply not using A -> UNDER
\end{pbody}

\cardpart{Output contract}{canonical subtype labels; JSON only}
\begin{pbody}
Respond with a JSON object containing:
- "verdict": one of "RIGHT", "OVER\_PERSONALIZED", "UNDER\_PERSONALIZED"
- "subtypes": the primary error subtype if applicable, drawn from: under.complete\_omission, under.partial\_omission, under.underweighting, under.explicit\_violation, over.irrelevant\_use, over.scope\_leakage, over.stale\_or\_superseded, over.unsupported\_inference, over.over\_dominance; an empty list if RIGHT
- "confidence": float 0.0--1.0
- "reasoning": brief explanation citing which preference was mishandled and how
Respond only with the JSON object, no other text.
\end{pbody}
\end{promptcard}

\begin{promptcard}{Evidence-first evaluation}{prompt:evfirst}
\cardmeta{Purpose}{two-stage elicitation: extract candidate states first, then judge against the extraction}
\cardmeta{Input}{stage 1: context and request; stage 2: extracted evidence, request, and response}
\cardmeta{Output}{stage 2 emits the same JSON contract as Prompt~\ref{prompt:e2e}}

\cardpart{Stage 1: evidence extraction}{surfaces candidate states with provenance and currency}
\begin{pbody}
Extract all user preferences, constraints, and states from this conversation context that are relevant to the user's current request.
Conversation context: \{context\}
User's current request: \{user\_request\}
List each preference, constraint, or state that could be relevant to answering the user's request. For each, note: what the preference/state is; which turn it was expressed in; whether it is still current or potentially stale/superseded.
Respond with a structured list of extracted evidence.
\end{pbody}

\cardpart{Stage 2: judgment}{same taxonomy, disambiguation rule, and output contract as Prompt~\ref{prompt:e2e}, applied to the extracted evidence}
\begin{pbody}
Given the extracted evidence below, evaluate whether the assistant response correctly uses these preferences, constraints, and states.
Extracted evidence: \{evidence\}
User request: \{user\_request\}
Assistant response: \{response\}
[Verdict taxonomy, disambiguation rule, and JSON output contract identical to Prompt~\ref{prompt:e2e}.]
\end{pbody}
\end{promptcard}

\begin{promptcard}{State-table evaluation}{prompt:agentic}
\cardmeta{Purpose}{structured elicitation: search the dialogue, build an explicit state table, then judge}
\cardmeta{Input}{target request, candidate response, searchable dialogue history}
\cardmeta{Output}{\ttfamily\footnotesize \{state\_table, verdict, subtypes, confidence, reasoning\}}

\cardpart{Role and inputs}{frames evaluation as systematic state reconstruction}
\begin{pbody}
You are a persona-use evaluator. You systematically search the dialogue history, build a structured state table, and then judge the response.
Target query: \{user\_request\}
Candidate response: \{response\}
Searchable multi-episode dialogue history: \{context\}
\end{pbody}

\cardpart{Evidence search}{bounded search over candidate states}
\begin{pbody}
Step 1: List which user preferences, constraints, or states could be relevant to the target query. Consider: dietary, budget, scheduling, location, people involved, prior decisions, hard constraints, soft preferences.
Step 2: For each candidate state, search the dialogue (at most 6 searches) for supporting evidence, later updates or exceptions, conflicting evidence, and holder/scope boundaries (does it belong to this user, for this task, at this time?).
\end{pbody}

\cardpart{State-table schema}{the explicit applicability record per state}
\begin{pbody}
Step 3: For each confirmed state, record: State | Holder | Scope | Status | Strength | Evidence Turns | Policy, where
- Holder: user / spouse / parent / colleague / other
- Scope: global / specific domain / specific task / specific time
- Status: active / updated / superseded / expired / one-time exception
- Strength: hard (dealbreaker) / strong (important) / soft (nice-to-have)
- Evidence Turns: the turn numbers that support it
- Policy: IGNORE (not applicable here) / SUPPORT (should inform but not dominate) / DOMINATE (must be satisfied)
\end{pbody}

\cardpart{Response checks}{maps table-policy violations to verdicts}
\begin{pbody}
Step 4: Check that the response (1) uses all states with Policy=DOMINATE (missing any -> UNDER); (2) appropriately references SUPPORT states (ignoring all -> UNDER, over-applying -> OVER); (3) does not use IGNORE states (using them -> OVER); (4) does not treat DOMINATE states as merely SUPPORT (-> UNDER, underweighting); (5) does not treat SUPPORT states as DOMINATE (-> OVER, over-dominance); (6) if it contradicts State A by using old/superseded information -> OVER (stale), not UNDER; (7) if it contradicts State A by simply forgetting it (no old information referenced) -> UNDER (explicit violation).
Step 5: Reach a verdict.
\end{pbody}

\cardpart{Constraints}{guards against assuming an error and against multi-error verdicts}
\begin{pbody}
- Do not assume the response contains an error; many responses are correctly personalized.
- Use at most 6 searches through the dialogue.
- Return only one primary error when the verdict is UNDER or OVER.
\end{pbody}

\cardpart{Output contract}{state table plus the standard verdict fields}
\begin{pbody}
Respond only with a JSON object:
\{"state\_table": [\{"state": "...", "holder": "...", "scope": "...", "status": "...", "strength": "...", "evidence\_turns": [...], "policy": "..."\}], "verdict": "RIGHT" or "OVER\_PERSONALIZED" or "UNDER\_PERSONALIZED", "subtypes": [the primary error subtype, drawn from the canonical list of Prompt~\ref{prompt:e2e}; empty if RIGHT], "confidence": 0.0--1.0, "reasoning": "concise justification"\}
\end{pbody}
\end{promptcard}

\end{document}